\documentclass[a4paper,fleqn]{article}
\usepackage{a4wide}
\usepackage{amsmath,amssymb,amsthm}
\usepackage{thmtools}
\usepackage{hyperref}
\usepackage[utf8]{inputenc}
\usepackage[T1]{fontenc}
\usepackage{xifthen} 
\usepackage{xspace}
\usepackage{booktabs}
\usepackage{multirow}
\usepackage{graphicx}
\usepackage{colortbl}
\usepackage{longtable}
\usepackage{wrapfig}
\usepackage{rotating}
\usepackage{enumerate}
\usepackage{xcolor}	

\usepackage[noend]{algpseudocode}
\usepackage{algorithm,algorithmicx}
\MakeRobust{\Call}

\newcommand{\signal}[2][]{#2\ifthenelse{\isempty{#1}}{}{(#1)}}
\newcommand{\sut}{\mathcal{M}}
\newcommand{\inputspace}{\mathcal{I}}
\newcommand{\outputspace}{\mathcal{O}}
\newcommand{\generator}{\mathcal{G}}

\newcommand{\similarity}{f}
\newcommand{\diversity}{\mathcal{D}}
\newcommand{\divloss}{L}
\newcommand{\always}{\square}

\newcommand{\R}{\mathbb{R}}

\newcommand{\critical}{\mathcal{C}}
\newcommand{\distribution}{\mathcal{P}_\theta}
\newcommand{\target}{\mathcal{P}_r}

\newcommand{\prob}[1]{\mathbb{P}(#1)}

\newcommand{\alg}{\mathcal{A}}
\newcommand{\latentspace}{\mathcal{H}}

\newcommand{\eventually}{\lozenge}

\newcommand{\best}{\cellcolor{green!25}}

\newcommand{\worst}{\cellcolor{orange!25}}
\newcommand{\bestbox}{{\fboxsep=0pt\fbox{\color{green!25}\rule{3mm}{3mm}}}\hphantom{a}}
\newcommand{\secondbox}{{\fboxsep=0pt\fbox{\color{white}\rule{3mm}{3mm}}}\hphantom{a}}
\newcommand{\thirdbox}{{\fboxsep=0pt\fbox{\color{blue!25}\rule{3mm}{3mm}}}\hphantom{a}}
\newcommand{\worstbox}{{\fboxsep=0pt\fbox{\color{orange!25}\rule{3mm}{3mm}}}\hphantom{a}}

\newcommand{\AFC}{\textup{AFC27}}
\newcommand{\ATI}{\textup{AT1}_{20}}

\newcommand{\ATVIA}{\textup{AT6}_{4,35,3000}}
\newcommand{\ATVIB}{\textup{AT6}_{8,50,3000}}
\newcommand{\ATVIC}{\textup{AT6}_{20,65,3000}}

\newcommand{\CCIII}{\textup{CC3}}
\newcommand{\CCIV}{\textup{CC4}}
\newcommand{\FA}{\textup{F16}}
\newcommand{\NN}{\textup{NN}}
\newcommand{\BEAMNG}{\textup{BEAMNG}}
\newcommand{\AMBIEGEN}{\textup{AMBIEGEN}}

\title{Learning test generators for cyber-physical systems}
\author{
Jarkko Peltomäki and Ivan Porres\\
Faculty of Science and Engineering, \\ Åbo Akademi University \\ Turku, Finland \\
\texttt{name.surname@abo.fi}
}
\date{}

\begin{document}
\maketitle

\begin{abstract}
Black-box runtime verification methods for cyber-physical systems can be used to discover errors in systems whose inputs and outputs are expressed as signals over time and their correctness requirements are specified in a temporal logic.  Existing methods, such as requirement falsification, often focus on finding a single input that is a counterexample to system correctness.
In this paper, we study how to create test generators that can produce multiple and diverse counterexamples for a single requirement. Several counterexamples expose system failures in varying input conditions and support the root cause analysis of the faults.

We present the WOGAN algorithm to create such test generators automatically. The algorithm works by training iteratively a Wasserstein generative adversarial network that models the target distribution of the uniform distribution on the set of counterexamples. WOGAN is an algorithm that trains generative models that act as test generators for runtime verification. The training is performed online without the need for a previous model or dataset. We also propose criteria to evaluate such test generators. 

We evaluate the trained generators on several well-known problems including the ARCH-COMP falsification benchmarks. Our experimental results indicate that generators trained by the WOGAN algorithm are as effective as state-of-the-art requirement falsification algorithms while producing tests that are as diverse as a sample from uniform random sampling.
We conclude that WOGAN is a viable method to produce test generators automatically and that these test generators can generate multiple and diverse counterexamples for the runtime verification of cyber-physical systems.
\end{abstract}

\section{Introduction}
Cyber-physical systems (CPSs)~\cite{2011:a_survey_of_cyber_physical_systems} integrate heterogeneous hardware and software components, and they interact autonomously with the physical environment in real time. Examples CPSs range from medical monitoring systems to autonomous vehicles. Due to the interaction with the physical environment, CPSs may cause injuries, environmental damage, or economic losses if they fail~\cite{2002:safety_critical_sytems_challenges_and_directions}. Thus it is critical to assess the correctness of a CPS before it is operated.

Requirement falsification \cite{DBLP:journals/jair/CorsoMKLK21} is an approach to CPS validation that searches for system inputs that yield outputs that violate a single given system requirement. A requirement falsification algorithm iteratively generates an input, executes it on the system, observes the system behavior with a runtime monitor, and evaluates if the requirement is satisfied or not. If a falsifying test not conforming to the requirement is found, then the test is a counterexample against the claim that the system is safe with respect to the requirement. Due to the real-time nature of CPS, execution traces of CPSs are typically represented as signals over time and the requirements are expressed in signal temporal logic (STL) \cite{maler2004monitoring}. A popular approach to requirement falsification is to turn the task of finding a falsifying test into an optimization problem via a robustness metric. A robustness metric is a function from a system trace to a real number. The metric takes a negative value if the trace does not satisfy the requirement. If the requirement is satisfied by the trace, then the robustness is positive and indicates how close the trace is being falsifying. See, e.g.,  \cite{2011:staliro_a_tool_for_temporal_logic_falsification,2020:approximation_refinement_testing_of_compute_intensive,2020:falsification_of_cyber_physical_systems_with_robustness,2021:falsification_of_hybrid_systems_using_adaptive,2021:stochastic_optimization_with_adaptive_restart_a_framework,2021:psy_taliro_a_python_toolbox_for_search,2023:search_based_software_testing_driven,2024:falsification_using_reachability_of_surrogate_koopman,ogan_full} for tools and algorithms that approach requirement falsification using a robustness metric.

Traditionally, requirement falsification algorithms aim to produce a single falsifying test as this is enough to falsify a requirement. This is, for example, the goal of the ARCH-COMP competition \cite{ARCH23} for the validation of CPSs with STL requirements. However, it is useful to search for several falsifying tests. Several different violations of a requirement can aid in determining the root cause of the observed faults saving developer time in correcting the system. Moreover, even when a system is safe, several tests that trigger extremal behavior are useful in understanding the system better in extreme conditions. This is especially important if the tests are found in simulated conditions but are yet to be tested in actual field operations. Finding several falsifying tests is particularly significant in scenario-based testing when creating scenarios manually is labor-intensive. This applies in particular to the testing of autonomous vehicles as it is of paramount importance to test such vehicles in varied traffic and environmental conditions. For this very reason recent research on testing of autonomous has focused on finding test suites of falsifying tests that are diverse \cite{2024:diversity_guided_search_exploration_for_self,SBFT-toolcomp24,2022:a_search_based_framework_for_automatic_generation,2021:deephyperion_exploring_the_feature_space_of_deep,2021:salvo_automated_generation_of_diversified_tests,2020:model_based_exploration_of_the_frontier_of_behaviours}.

Test suite diversity is a broad theme, but informally a diverse test suite exercises the system broadly or covers its behavior extensively. The concept of diversity also appears in other contexts than requirement falsification. For example, in software testing, test suites are often expected to have high code coverage \cite{2017:introduction_to_software_testing}. This sort of diversity is not about finding diverse faults per se but about observing the system in various circumstances. Diversity is often ensured by selecting tests for dissimilarity. Similarity of tests is measured by a similarity metric based on some aspects of a test such as its structure or the behavior of the system once the test is executed \cite{2023:a_survey_of_the_metrics_uses_and_subjects_of_diversity}.

In this article, by \emph{input diversity}, we mean test suite diversity measured in terms of the input structure without regard to the system behavior. For example, creating geometrically different roads for an autonomous vehicle to drive on ensures input diversity \cite{2021:salvo_automated_generation_of_diversified_tests}. Sometimes input diversity can mean that tests are spread evenly in the input space \cite{2004:adaptive_random_testing}. Another type of diversity is \emph{behavioral diversity}, which also takes into account the system states and/or output. An example of behavioral diversity is requiring that an autonomous car takes different steering actions when all tests in a diverse test suite are executed \cite{2024:testing_cyber_physical_systems_with_explicit,2021:deephyperion_exploring_the_feature_space_of_deep,SBFT-toolcomp24}.

The focus of this article is input diversity under the condition of falsification, that is, we aim to create test suites of falsifying tests that have high input diversity. This form of diversity, which we dub \emph{IF-diversity}, is a combination of input diversity and behavioral diversity, and it has recently been considered in testing of autonomous vehicles \cite{2024:diversity_guided_search_exploration_for_self,SBFT-toolcomp24,2022:a_search_based_framework_for_automatic_generation,2021:deephyperion_exploring_the_feature_space_of_deep,2020:model_based_exploration_of_the_frontier_of_behaviours}. This recent research approaches the generation of IF-diverse test suites mainly through evolutionary algorithms that attempt to create an IF-diverse population of falsifying tests using one or more objective functions that promote falsification and input diversity.

In this article, we identify a new problem related to that of creating an IF-diverse test suite. Given a system and a requirement with an associated robustness metric, we consider the requirement falsification generator problem (RFG-problem) of training a generator model that can sample from the uniform distribution on falsifying tests for the system. Given such a generator, creating an IF-diverse test suite is just a matter of sampling the generator. Our main contribution is the online Wasserstein generative adversarial (WOGAN) algorithm that trains a generator that approximates a solution of the RFG-problem, that is, we train a generative model that models the set of falsifying tests and propose a sampling method for sampling a test suite with the generator. The WOGAN algorithm models the generator as a Wasserstein generative adversarial network (WGAN), which is known to be able to generate more diverse samples from its target distribution than other generative modeling techniques such as generative adversarial networks \cite{wgan,wgan_gp}. Since the generator trained by WOGAN is only an approximation, it is not guaranteed that it can be used to sample a falsifying test when the requirement is falsifiable. However, the training of the generator is set up in such a way that we expect to sample only tests with low positive robustness.

A major difference to previous research on IF-diverse test suites is that we do not just produce a static test suite but a model for the set of low-robustness tests. This means that we can sample more tests at will any time after the generator training has finished. If the generator was trained in simulated conditions, the generator can be used to generate test suites for the system under field conditions. In addition, the trained generator can be retrained. Either it can be retrained on the same system and requirement to improve the model and the consequent test suites, or it can be used on an altered system and an altered requirement via transfer learning. This latter point is important as altering a system slightly can invalidate a test suite but retraining a model to match an altered system is possibly more efficient than training a model from scratch.

The WOGAN algorithm can be used to solve the RFG-problem for any deterministic black-box system with outputs modeled as signals over time and any requirement with an associated robustness metric. Thus WOGAN is applicable in cases where the requirement is expressed as an STL formula. In particular, WOGAN can be used for the CPS benchmarks of the ARCH-COMP competition. A important feature of WOGAN is that it works tabula rasa: no prior data or models are required; WOGAN generates all required training data online during its execution.

As the RFG-problem has not been identified prior to this work in the context of CPS validation, we present in this article criteria according to which a generator can be evaluated to see how well it approximates a solution to the RFG-problem. We evaluate the generators trained by the WOGAN algorithm according to the following two criteria: i) the input diversity of a generator sample measured according to a test similarity measure, ii) the shape of the robustness distribution of a generator sample measured by computing certain quantiles of the distribution. We consider a generator a good approximation to a solution to the RFG-problem if its robustness distribution generally takes low values and if the tests sampled by it are highly input-diverse. We compare the WOGAN generators against uniform random generators on several standard CPS falsification benchmarks derived from the ARCH-COMP 2023 competition \cite{ARCH23} and the SBFT 2023 competition \cite{SBFT-toolcomp23}. The benchmarks of the ARCH-COMP competition range from testing the automatic transmission of a car to testing the ground collision avoidance system of a fighter jet. The CPS benchmark of SBFT 2023 concerns the lane keeping assist system of an autonomous car. Our experiment results show that WOGAN generator samples are typically as diverse as uniform random samples and that the corresponding robustness distribution is much lower than the robustness distributions for uniform random samples. We also evaluate the falsification capability of WOGAN generators against state-of-the-art falsification tools from the ARCH-COMP 2023 competition and find that the  generators trained by WOGAN typically compare favorably against the state-of-the-art tools in terms of falsification effectiveness.

Variants of the WOGAN algorithm have previously been used in the SBST 2022 \cite{SBST-toolcomp22,sbst_long,sbst_short}, SBFT 2023 \cite{SBFT-toolcomp23,wogan_sbft23}, and SBFT 2024 \cite{SBFT-toolcomp24-uav,wogan_uav} competitions. In the former two competitions, WOGAN was used to test an autonomous car in a simulator, and WOGAN was among the top tools in the final tool ranking. In the SBFT 2024 competition, it was used to test drone flight paths, and WOGAN was ranked as the best algorithm to do so. It is worth noticing that WOGAN was applicable to all competitions since the generator training uses no domain knowledge. The main difference to the current article is that in these competitions WOGAN's performance was evaluated based on the tests it executed during generator training meaning that the WOGAN generators, the main topic of the current article, were not discussed at all. Moreover, the WOGAN implementation is only briefly discussed in the paper \cite{sbst_long}, and this paper does not give the full details, which are now present here for the first time. WOGAN has also been used in the paper \cite{2024:testing_cyber_physical_systems_with_explicit} which is about finding test suites for a self-driving car that are behaviorally diverse. This work focuses on falsifying several explicitly stated STL requirements using several parallelly executing WOGAN algorithms. On the contrary, in the current article, we focus only on falsifying a single requirement.

The main contributions of this article, not present in our previous research, are:
\begin{itemize}
    \item The identification of the RFG-problem of sampling diversely the set of falsifying tests of a system requirement with an associated robustness metric.
    \item The novel WOGAN algorithm based on training a Wasserstein generative adversarial network for approximating the solution of the RFG-problem. The algorithm applies to deterministic black-box systems, supports arbitrary STL requirements, and does not require any previous model or dataset for the system. The proposed implementation and hyperparameter setup provides competitive results on a wide range of benchmarks.
    \item A novel methodology to evaluate a generator to see how well it approximates the solution of the RFG-problem.
    \item A comprehensive evaluation of the WOGAN generators against uniform random sampling and state-of-the-art CPS falsification tools. The evaluation includes evaluating the effect of various design choices in the WOGAN algorithm.
\end{itemize}

We proceed as follows. \autoref{sec:problem_description} gives the necessary background information on STL and describes the RFG-problem and related work. \autoref{sec:wgan} briefly discusses Wasserstein generative adversarial networks and their training as necessary background to the WOGAN algorithm, which is fully described in \autoref{sec:wogan}. \autoref{sec:research_questions} states our research questions, and the methodology to answer them is developed in \autoref{sec:evaluation_methodology}. The benchmarks to be used in the experiments to answer the research question and the WOGAN implementation details are explained in Sections~\ref{sec:benchmark_selection},\ref{sec:wogan_setup}, and the actual experiment results and their analysis are found it \autoref{sec:results}. We end the paper with the conclusions of \autoref{sec:conclusions}.

\section{Problem Description and Related Work}\label{sec:problem_description}

\subsection{SUT}
A \emph{system under test} (SUT) is a deterministic function $\sut$ from an \emph{input space} $\inputspace$ to an \emph{output space} $\outputspace$. We assume that the spaces $\inputspace$ and $\outputspace$ are either compact subsets of a Euclidean vector space or function spaces containing signals $[0, T] \to \R^n$ for some $T > 0$. The elements $(t, \sut(t))$, $t \in \inputspace$, are called \emph{traces}. For brevity, we often write $\sut(t)$ in place of $(t, \sut(t))$.

In the automatic transmission benchmark, the inputs are two signals $s_1\colon [0, 30] \to [0,100]$, $s_2\colon [0,30] \to [0,325]$ representing the vehicle's throttle and brake. The output signals are the speed of the vehicle, the engine speed of the vehicle, and the current gear given as signals over the time domain $[0,30]$. In the ground collision avoidance benchmark, the input is a vector in $[0.2\pi, 0.2833\pi] \times [-0.4\pi, -0.35\pi] \times [-0.375\pi, -0.125\pi]$ representing the initial roll, pitch, and yaw of the jet. The output is a signal describing the altitude of the jet during the $15 \textup{s}$ simulation.

\subsection{System Requirements}
We assume that the requirement for the SUT is given as an STL formula. An example STL formula is
$\always_{t \in [0,30]} \, \signal[t]{v} < 120$, which expresses the requirement that during the first $30$ time units the signal $\signal[t]{v}$ should always be at most $120$. We do not define STL precisely here. Instead we refer the reader to \cite{maler2004monitoring}. With the notation $(s, t) \models \varphi$, we mean that the trace $s$ satisfies the STL formula $\varphi$ when evaluated at time $t$. On the contrary, we take $(s, t) \not\models \varphi$ to indicate that $\varphi$ is not satisfied and the trace $s$ evaluated at time $t$ is a counterexample to $\varphi$.

Associated to an STL formula $\varphi$, it is possible to find a robustness metric $\rho$ which maps system traces (evaluated at time $t$) to real numbers. The map $\rho$ has the properties
\begin{align*}
    \rho(s, t) > 0 &\text{ if } (s, t) \models \varphi,  \\
    \rho(s, t) < 0 &\text{ if } (s, t) \not\models \varphi.
\end{align*}
The case when $\rho(s, t) = 0$ is ambiguous. This does matter not for the purposes of this article, and we can view $\varphi$ to be falsified in this case. In addition to the preceding properties, the closer $\rho(s, t)$ is in absolute value, the smaller the change is needed to perturb the trace $s$ for $\rho(s, t)$ to change sign; see \cite[Sect.~3.4]{2009:robustness_of_temporal_logic_specifications_for_continuous} for the precise statement. The properties of $\rho$ enable the search for a falsifying test to be turned into the problem of minimizing $\rho$. The robustness metric $\rho$ can be effectively computed given $\varphi$, and $\rho$ is efficient to evaluate. For the full details, see \cite{donze2010robust}.

Machine learning algorithms often perform better with normalized data \cite{2012:efficient_backprop}. For this reason, we work with a scaled robustness metric $\overline{\rho}$ taking values in the interval $[0,1]$, which is defined by setting
\begin{equation*}
    \overline{\rho}(s, t) =
    \begin{cases}
        0 & \text{ if $\rho(s, t) \leq 0$}, \\
        \rho(s, t) / K(s, t) & \text{ if $\rho(s, t) > 0$,}
    \end{cases}
\end{equation*}
where $K(s, t)$ is an upper bound for the value of $\rho(s, t)$. The bound $K(s, t)$ is computed by finding the effective range for $\varphi$ given the trace $s$ as explained in \cite{ogan_full}. Computing the effective range requires ranges to be known for all system inputs and outputs. The scaled robustness metric $\overline{\rho}$ retains the property that falsifying $\varphi$ amounts to minimizing $\overline{\rho}$.

In this article, we evaluate the truth value of a requirement $\varphi$ always at time $t = 0$. This is not a restriction: all discussion is easily generalizable to cases where $t > 0$.

\subsection{The RF-problem}
Consider an SUT $\sut\colon \inputspace \to \outputspace$ and a requirement $\varphi$ with an associated scaled robustness metric $\rho\colon \inputspace \times \outputspace \to [0,1]$. The \emph{requirement falsification problem} (RF-problem) asks to falsify $\varphi$, that is, to produce an input $t$ in $\inputspace$ such that the corresponding system trace $\sut(t)$ is a counterexample to $\varphi$, that is, $(\sut(t), 0) \not\models \varphi$. Due to the properties of the function $\rho$, this amounts to the task
\begin{align*}
 \text{find $x \in \inputspace$ such that $\rho(\sut(t)) = 0$}.
\end{align*}
By an \emph{RF-algorithm}, we mean an algorithm that attempts to solve the RF-problem.

\subsection{The RFG-Problem}\label{ssec:rfg_problem}
Let $\sut$, $\varphi$, and $\rho$ be as in the previous subsection.
Let $\latentspace$ denote a \emph{latent space}, a subset of $\R^d$
for some $d$. A test generator, in short a \emph{generator}, is a mapping
$\generator\colon \latentspace \to \inputspace$. A generator $\generator$ is used to generate tests $\generator(x)$ by
sampling points $x$ of $\latentspace$ according to a probability distribution on $\latentspace$. In particular, a
generator $\generator$ is a random variable. As we are interested in falsifying tests in relation to the requirement
$\varphi$, we seek for generators $\generator$ such that the random variable
$\rho \circ \sut \circ \generator\colon \latentspace \to [0,1]$ tends to take small values. Let us next formalize this
further.

Let $\critical_\varepsilon$ be the \emph{set of $\varepsilon$-critical tests}, that is, set
$\critical_\varepsilon = \{t \in \inputspace : \rho(\sut(t)) \leq \varepsilon\}$. We say that a generator is
a \emph{$D_\varepsilon$-generator} if its distribution is the uniform distribution supported on $\critical_\varepsilon$
for some $\varepsilon$. A $D_\varepsilon$-generator has two main properties: i) it generates tests with robustness at
most $\varepsilon$ and ii) it is input-diverse in the sense that it generates varied tests in $\critical_\varepsilon$.
The latter property is important as it is easy to find nondiverse generators that generate tests in, e.g.,
$\critical_0$. Indeed, if $t$ in $\inputspace$ is a test with $\rho(\sut(t)) = 0$, then a constant generator that
always outputs $t$ has its distribution supported on a subset of $\critical_0$. However, with the exception of
degenerate systems with a single falsifying test, this constant generator does not satisfy ii). Thus the condition ii)
ensures that a $D_\varepsilon$-generator is not \emph{collapsed}: a collapsed generator generates very few distinct inputs and lacks
diversity. This so-called mode collapse is a common problem in generative adversarial network training \cite{2017:unrolled_generative_adversarial_networks}.

Recall that the requirement falsification problem (RF-problem) ask to find a single falsifying input. Motivated by
this problem and the above definitions, we define the \emph{requirement falsification generator problem} (RFG-problem)
as the problem of finding a $D_0$-generator given a SUT $\sut$ and a requirement $\varphi$ with an associated scaled robustness metric $\rho$.

The RFG-problem is different from the RF-problem in the sense that the latter can be easily verified as solved. Given a
candidate generator $\generator$ for the RFG-problem, we cannot easily prove that it is a $D_0$-generator. In practice,
we can only expect $\generator$ to be similar to a $D_0$-generator according to some computable metrics. We propose
two such metrics in \autoref{sec:evaluation_methodology}: a diversity score and quantile scores which respectively measure
the input diversity and capability of generating low-robustness tests. We use these measures to evaluate all generators in this paper.

\subsection{RFG-algorithms}
An \emph{RFG-algorithm} is an algorithm that solves the RFG-problem, that is, it outputs a generator $\generator$ as described in \autoref{ssec:rfg_problem}. The evaluation of the RFG-algorithm is based on a sample from $\generator$, which is evaluated according to preselected metrics to see how similar $\generator$ is to a $D_0$-generator. An example baseline RFG-algorithm is random search, which simply returns a generator that samples the input space $\inputspace$ of the SUT according to some probability distribution. In this article, we sample $\inputspace$ uniformly randomly.

In this article, we assume that an RFG-algorithm $\alg$ has no prior knowledge of the RFG-problem: $\alg$ gets as its inputs only the SUT $\sut$ and the robustness metric $\rho$. The algorithm $\alg$ is allocated an execution budget $B$ that limits the number of system trace $\sut(t)$ evaluations it is allowed to perform in order to create a generator.

\subsection{Related Work}\label{ssec:related_work}
On requirement falsification and RF-algorithms, we refer the reader to the ARCH-COMP competition \cite{ARCH21,ARCH22,ARCH23}. It suffices to say that tried approaches to requirement falsification include well-known metaheuristics such as simulated annealing or CMA-ES, Bayesian optimization, or evolutionary algorithms \cite{2024:reinforcement_learning_informed_evolutionary_search,ARCH23,2011:staliro_a_tool_for_temporal_logic_falsification,2021:stochastic_optimization_with_adaptive_restart_a_framework,2021:psy_taliro_a_python_toolbox_for_search}. Some approaches train a surrogate model or learn an automaton for the SUT and refine them according to SUT execution traces \cite{2020:approximation_refinement_testing_of_compute_intensive,2020:falsification_of_cyber_physical_systems_with_robustness}. We want to point out the OGAN RF-algorithm \cite{ogan_full} developed by the authors of the current article. This algorithm takes inspiration from generative adversarial networks but does not train a generator in the sense of \autoref{ssec:rfg_problem}. WOGAN uses as its WGAN sampling procedure the discriminator sampling of the OGAN algorithm.

The problem of finding multiple and diverse faults for a single requirement seems mainly been studied in relation to testing autonomous vehicles. This body of research includes the testing of the lane-keeping assist of the driving agent of the BeamNG.tech simulator \cite{SBFT-toolcomp24,2024:diversity_guided_search_exploration_for_self,SBFT-toolcomp23,SBST-toolcomp22,2022:a_search_based_framework_for_automatic_generation,SBST-toolcomp21,2021:deephyperion_exploring_the_feature_space_of_deep,2021:salvo_automated_generation_of_diversified_tests,2020:model_based_exploration_of_the_frontier_of_behaviours} as well as the flight paths of a drone \cite{SBFT-toolcomp24-uav}. The common theme is IF-diversity: falsification concerns misbehavior (vehicle out of lane or a drone crashing into an obstacle) and the faults are expected to be diverse as in that they cover a feature map extensively \cite{SBFT-toolcomp24-uav,SBFT-toolcomp24,SBFT-toolcomp23,SBST-toolcomp22,SBST-toolcomp21,
2021:deephyperion_exploring_the_feature_space_of_deep}. Notice that finding multiple faults for a single requirement is different from falsifying several requirements \cite{2021:efficient_optimization_based_falsification_of_cyber,2022:falsification_of_multiple_requirements_for_cyber_physical}.

Achieving diversity is often approached by maximizing a preselected dissimilarity metric. Either new tests are explicitly required to be dissimilar to existing tests or diversity is implicitly achieved by maximizing a dissimilarity objective. Used dissimilarity metrics are based on Euclidean distance \cite{2024:diversity_guided_search_exploration_for_self,2019:multi_objective_search_for_effective_testing,2020:model_based_exploration_of_the_frontier_of_behaviours}, Jaccard distance \cite{2022:a_search_based_framework_for_automatic_generation}, Levenshtein distance \cite{2020:model_based_exploration_of_the_frontier_of_behaviours,2022:a_search_based_framework_for_automatic_generation}, and feature maps \cite{2021:deephyperion_exploring_the_feature_space_of_deep} to name a few. For more dissimilarity metrics and further references, see \cite{2023:a_survey_of_the_metrics_uses_and_subjects_of_diversity}. In the current paper, we do not explicitly select tests for dissimilarity and neither do we maximize an objective relating to diversity. The advantage of this is that there is no need to define a domain-specific diversity metric. For us, diversity is implicitly defined by the WGAN, which is capable of producing diverse tests from its target distribution \cite{wgan,wgan_gp}. Our evaluation of input diversity, however, is explicitly based on dissimilarity metrics; see \autoref{ssec:diversity}.

\section{Wasserstein Generative Adversarial Networks}\label{sec:wgan}
In this section, we briefly introduce the concept of a Wasserstein Generative Adversarial Network (WGAN) and describe how it is trained. The full details are found in \cite{wgan,wgan_gp}.

Let $\latentspace$ be a latent space as in \autoref{ssec:rfg_problem} and $\inputspace$ to be the input space of a SUT. Consider a parametric generator $\generator_\theta\colon \latentspace \to \inputspace$ with parameters $\theta$. The goal is to select $\theta$ in such a way that the distribution $\distribution$ of $\generator_\theta$ matches a target distribution $\target$ on $\inputspace$. This means that when $\latentspace$ is sampled, say, uniformly randomly and the resulting samples are mapped via $\generator_\theta$, the resulting samples follows the distribution $\target$. When $\generator_\theta$ is trained as a WGAN, this is achieved by minimizing the Wasserstein distance between the distributions $\distribution$ and $\target$.

The Wasserstein distance between the distributions $\distribution$ and $\target$ is given by
\begin{equation}\label{eq:wd}
    \inf_{\gamma \in \Pi(\distribution, \target)} \mathbb{E}_{(x,y) \sim \gamma} \left( ||x - y|| \right),
\end{equation}
where $\Pi(\distribution, \target)$ is the set of all joint distributions $\gamma(x, y)$ with marginals $\distribution$ and $\target$. Intuitively speaking, a joint distribution $\gamma(x, y)$ indicates how much probability mass must be transported from $x$ to $y$ to transform $\distribution$ to $\target$. The Wasserstein distance is the cost of the optimal transport.

The formula \eqref{eq:wd} is not directly usable as an objective function, but the Kantorovich-Rubinstein duality shows that the distance can be expressed as
\begin{equation}\label{eq:dual}
    \sup_{||f||_L \leq 1} \left( \mathbb{E}_{x \sim \target}(f(x)) - \mathbb{E}_{x \sim \distribution}(f(x)) \right),
\end{equation}
where the supremum is over all $1$-Lipschitz functions $f\colon \inputspace \to \R$. This dual formulation suggests that the training can be accomplished if the $1$-Lipschitz function $f$, too, is learned. It is proved in \cite[Thm.~3]{wgan} that this is possible: under mild assumptions, there exists a solution $f$ for the problem 
\begin{equation}\label{eq:wgan_objective}
    \max_{||f||_L \leq 1} \left( \mathbb{E}_{x \sim \target}(f(x)) - \mathbb{E}_{x \sim \distribution}(f(x)) \right),
\end{equation}
and the solution can be found via gradient descent. The solution $f$, written henceforth as $f_w$ with parameters $w$, is called the \emph{critic}.

The resulting training algorithm is summarized in \autoref{alg_wgan}. The training assumes that we have access to some training data in that we have samples from the target distribution $\target$. The lines $2$--$11$ are responsible for training the critic when $\generator_\theta$ is fixed. The first part of the critic loss function on line $9$ corresponds to the objective \eqref{eq:wgan_objective}, but it also has an additional term. The purpose of this term is to softly enforce the condition that $f_w$ is supposed to be a $1$-Lipschitz function. For the full details, see \cite{wgan_gp}. After the critic has been trained, the generator is updated to minimize the Wasserstein distance. Indeed, the gradient on line $13$ is the gradient (with respect to $\theta$) of the Monte-Carlo estimator corresponding \eqref{eq:dual}. In \cite{wgan_gp} the values $\lambda = 10$ and $n_{\textup{critic}} = 5$ are suggested.

\begin{algorithm}[t]
  \caption{Algorithm to train a WGAN.}\label{alg_wgan}
  \begin{algorithmic}[1]
    \Require{The number $n_{\textup{critic}}$ of critic updates per generator update, the batch size $m$, and the gradient penalty coefficient $\lambda$.}
    \Statex
    \While{$\theta$ has not converged}
        \For{$t \in \{1, \ldots, n_{\textup{critic}}\}$}
            \For{$i \in \{1, \ldots, m\}$}
                \State Sample a real data point $x \sim \target$.
                \State Sample a latent variable $z$ uniformly randomly from $\latentspace$.
                \State Sample a random number $\epsilon$ uniformly randomly from $[0,1]$.
                \State $\tilde{x} \gets \generator_\theta(z)$
                \State $\hat{x} \gets \epsilon x + (1 - \epsilon) \tilde{x}$
                \State $L_i \gets f_w(\tilde{x}) - f_w(x) + \lambda(||\nabla_{\hat{x}} f_w(\hat{x})||_2 - 1)^2$
            \EndFor
            \State $g_w \gets \nabla_w \frac{1}{m} \sum_{i=1}^m L_i$
            \State Update critic parameters $w$ according to the gradient $g_w$.
        \EndFor
        \State Sample a batch $\{z_1, \ldots, z_m\}$ uniformly from $\latentspace$.
        \State $g_\theta \gets - \nabla_\theta \frac{1}{m} \sum_{i=1}^m f_w(\generator_\theta(z_i))$
        \State Update generator parameters $\theta$ according to the gradient $g_\theta$.
    \EndWhile
    \end{algorithmic}
\end{algorithm}

\section{The WOGAN Algorithm}\label{sec:wogan}
The WOGAN RFG-algorithm has first been described in \cite{sbst_long} and used previously in \cite{sbst_short,wogan_sbft23,2024:testing_cyber_physical_systems_with_explicit,wogan_uav}. Here we give for the first time a description of it with full details and motivations. We remark that the algorithm described here deviates slightly from the algorithm of \cite{sbst_long}; the deviation was necessary as the algorithm was applied to use cases beyond the original use case of \cite{sbst_long}. The pseudocode for the main loop of the WOGAN algorithm is found in \autoref{alg1}. Let us first go through the main phases of the algorithm before describing everything in detail.

\begin{algorithm}[t]
  \caption{Main loop of the WOGAN RFG-algorithm.}\label{alg1}
  \begin{algorithmic}[1]
    \Require{SUT $\sut$, execution budget $B$, random execution budget $B_R$, latent space dimension $d_H$, exploration probability $\theta$, training delay $d_T$.}
    \Statex
    \State $T \gets$ \Call{initialize\_test\_repository}{$\inputspace$}
    \State $W \gets$ \Call{initialize\_wgan}{$\inputspace, d_H$}
    \State $U \gets$ \Call{initialize\_uniform\_random\_sampler}{$\inputspace$}
    \State $A \gets$ \Call{initialize\_analyzer}{$\inputspace$}
    \State $S \gets$ \Call{initialize\_training\_data\_sampler}{\! }
    \State last\_trained $\gets 0$
    \While{$|T| < B$}
        \If{$|T| < B_R$ or \Call{random}{\! } $\leq \theta$}
            \State test  $\gets$ \Call{sample}{$U$}
        \Else
            \If{$|T| -$ last\_trained $\geq d_T$}
                \State \Call{train\_analyzer}{$A$; $T$}
                \State \Call{train\_wgan\_online}{$W$; $S$; $T$, $\{T[-1 - d_T + 1], \ldots, T[-1]\}$; $(B - |T|)/B_R$}
                \State last\_trained $\gets |T|$
            \EndIf
            \State test  $\gets$ \Call{rejection\_sample}{$W$; $A$}
        \EndIf

        \State robustness $\gets$ \Call{execute}{$\sut;$ test}
        \State \Call{record}{$T;$ test, robustness}
    \EndWhile
    \end{algorithmic}
\end{algorithm}

The aim of the algorithm is to train a WGAN $W$ with target distribution supported on the set $C_0$ of falsifying tests. As we do not assume any prior training data from $C_0$, we cannot directly apply \autoref{alg_wgan}. Instead, we alter the training data distribution during the algorithm execution. Initially, the WGAN is trained on low-robustness tests obtained via uniform random sampling. Later, we sample the trained WGAN itself to produce more low-robustness tests to further shift the training data distribution towards a distribution supported on $C_0$.

Lines $1$--$5$ of \autoref{alg1} initialize the main objects of the WOGAN algorithm. The test repository $T$ stores the tests executed on the SUT together with the associated robustness values. During each round of the main loop, starting on line $7$, a generator is selected to be sampled for a test (lines $8$--$15$), the sampled test is executed to learn its robustness (line $16$), and the test repository is updated (line $17$). During the first $B_R$ rounds, the selected generator is $U$ which samples tests uniformly randomly. After this initial random sampling phase is over, we deem that the test repository has enough data to train the WGAN, and the algorithm switches to train the analyzer $A$ (described in \autoref{ssec:analyzer_training}) and to train and sample the WGAN (lines $11$--$15$). How often the training is performed is controlled by a training delay parameter $d_T$ (line $11$). The WGAN sampling is done using a rejection sampling with the analyzer described in \autoref{ssec:wgan_sampling}. The collection of training data from the test repository and training of the WGAN based on this data (line $13$) is done as described in \autoref{ssec:wgan_training_online}. On line $8$, we have an exploration feature, that is, we sample a test uniformly randomly with probability $\theta$ instead of sampling and training the WGAN. The reason for this is twofold. First, the WGAN can get stuck into a local minima, so introducing an independent source of tests can help it get unstuck. Secondly, if the total execution budget is large, instead of setting $B_R$ large, we can set $B_R$ low and $\theta$ high to begin the WGAN training earlier. The probability $\theta$ can safely be set to $0$. The algorithm terminates when the SUT execution budget is exhausted.

\subsection{The Analyzer and Its Training}\label{ssec:analyzer_training}
The analyzer $A$ estimates the robustness metric $\rho\colon \inputspace \to [0, 1]$. This regression task can be approached by any suitable regression model, but here we choose to model $A$ as a neural network. Due to the assumption that we have little training data, we do one gradient update per epoch based on the single batch of training data consisting the whole test repository $T$ at the training time. The neural network structure, loss function, optimizer, learning rate, number of epochs, and other such details for the experiments of \autoref{sec:results} are described in \autoref{sec:wogan_setup}.

\subsection{WGAN Rejection Sampling}\label{ssec:wgan_sampling}
Here we describe line $15$ of \autoref{alg1}, that is, how to sample a WGAN for a new test to be executed on the SUT. We propose a form of rejection sampling with the help of the analyzer $A$. The idea is to sample the WGAN as described in \autoref{ssec:rfg_problem} but to reject a sample if $A$ estimates its robustness to be too high. As both the WGAN and the analyzer are imperfect, care needs to be taken to ensure that a sample is eventually obtained. To this end, we maintain an acceptance threshold. Any test estimated to have robustness below the threshold is accepted. If a test is not accepted, the threshold is increased. We use an initial acceptance threshold of $0$, and we increase it towards $1$ in a way analogous to repeatedly multiplying the number $1$ with $\alpha$, $0 < \alpha < 1$, to approach $0$. Our rejection sampling algorithm is described in \autoref{alg_rejection_sampler}. Line $10$ ensures loop termination if $A$ always estimates the robustness to be exactly $1$. We use a priority queue to store all sampled tests so that we can reconsider the test with the best estimated robustness so far once the threshold changes. We remark that the exactly same rejection sampling approach is used in \cite{ogan,ogan_full}. In this article, we set $\alpha = 0.95$.

\begin{algorithm}[t]
  \caption{Rejection sampling of a WGAN.}\label{alg_rejection_sampler}
  \begin{algorithmic}[1]
    \Require{Threshold increase multiplier $\alpha$.}
    \Statex
    \Function{rejection\_sample}{$W$; $A$}
    \State $P \gets$ \Call{initialize\_priority\_queue}{\! }
    \State accept\_threshold $\gets 0$
    \State $\epsilon \gets 0.0001$
    \Repeat
        \State candidate\_test $\gets$ \Call{sample}{$W$}
        \State estimated\_robustness $\gets$ \Call{estimate\_robustness}{candidate\_test; $A$}
        \State \Call{add\_to\_queue}{$P$; estimated\_robustness, candidate\_test}
        \State accept\_threshold $\gets 1 - \alpha \cdot (1 - \, $accept\_threshold$)$
    \Until{\Call{min}{$P$} $- \, \epsilon$ $\leq$ accept\_threshold}
    \State test $\gets$ \Call{test\_with\_min\_robustness}{$P$}
    \State \Return test
    \EndFunction
\end{algorithmic}
\end{algorithm}

\subsection{Online WGAN Training}\label{ssec:wgan_training_online}
The function TRAIN\_WGAN\_ONLINE described in \autoref{alg_sampler} is simple: for each WGAN training epoch, sample a batch $X$ of tests from the test repository $T$ with the training data sampler $S$ and train the WGAN with $X$. When $X$ is given, the WGAN is trained using \autoref{alg_wgan} ($X$ corresponds the batch of real data sampled on lines $3$-$9$ of \autoref{alg_wgan}). The models for the generator and the critic and their training hyperparameters for the experiments of \autoref{sec:results} are described in \autoref{sec:wogan_setup}.

If $T$ is static and $S$ samples $T$ uniformly randomly, the function TRAIN\_WGAN\_ONLINE is equivalent to the training of a WGAN using just \autoref{alg_wgan} with the tests of $T$ as training data. Since we do not initially expect the tests of $T$ to be a sample from our target distribution of falsifying tests, we need to sample a subset of $T$ and alter the sampling procedure over the execution of the WOGAN algorithm. The purpose of the sampler $S$ is to implement this functionality.

Our proposal for a sampler is given in the function QUANTILE\_SAMPLER of \autoref{alg_sampler}. The core idea is that only those tests of $T$ that form the lower proportion of size $Q$ when the tests are ordered according to to their robustness values are candidate to be included in the WGAN training data. The number $Q$ decreases during the WOGAN algorithm execution in such a way that initially many high-robustness tests can be candidates but this is no longer true at the end of the algorithm execution. This achieves the shift of the training data towards low-robustness tests over time. Since $Q$ is a proportional measure, the training data sampling is not directly tied to the maximum observed robustness value (which might vary greatly between different requirements).

On line $8$ of \autoref{alg_sampler}, the lower quantile $Q$ is computed based on the number $R$. The number $R$ represents how much of execution budget is left as a number in $[0, 1]$. When QUANTILE\_SAMPLER is called for the first time $R = 1$, and $R$ decreases linearly to $0$ as $|T|$ increases to $B$ (see line $13$ of \autoref{alg1}). We have made the decision that the function COMPUTE\_QUANTILE computes the function $0.4R + 0.1$, that is, $Q$ decreases linearly from $0.5$ to $0.1$ as $R$ decreases linearly from $1$ to $0$. 

After the quantile $Q$ has been determined, the tests in $T$ are binned on line $9$. This means that the tests that form the lower proportion of size $Q$ of $T$ are placed into $N_B$ bins. This is done by determining the maximum robustness $\rho_M$ of tests in the lower $Q$-quantile and placing a test with robustness $\rho$ into bin with index $i$ if $(i - 1)(\rho_M/N_B) \leq \rho < i(\rho_M/N_B)$ for some integer $i$ such that $1 \leq i \leq N_B$. The tests that do not belong to any bin are placed into the sink bin, which has index $N_B + 1$. In this article, we set $N_B = 10$.

After this, on line $10$, the bins receive weights for random sampling. The bin $i$ receives initially weight $N_B - i + 1$, and the sink bin has weight $0$. If a bin $i$ is empty (i.e., $|\textup{bins}[i]| = 0$), we set its weight to $0$. After this, the weights of the bins are scaled so that they sum up to $1$. Thus the lowest nonempty bin has most weight, and the highest bin has lowest weight.

After the bin weights have been determined, the actual sampling can begin. On lines $16$--$22$, a bin is selected using the bin weights as selection probabilities. If the selected bin happens to be empty (this is possible when several tests have already been included in the sample), the next bin is selected until a nonempty bin is found (the sink bin will contain tests). Then, on line $20$, a test is selected from the selected bin uniformly randomly, and the selected test is removed from the bin. Line $22$ adds the selected test to the training data sample.

The lines $11$--$15$ concern the test selection from the list NEW that includes the newest tests in $T$. On line $14$ of \autoref{alg1}, the list NEW is populated with the tests that have been sampled from the WGAN since the previous training of the analyzer and the WGAN (we assume that NEW contains always at least one test). The idea is to include a new test into the training data if it has low robustness. Without this feature, it is possible that the code on lines $16$--$22$ fails to include fresh tests in the training data. The condition to include a new test into the sample is given on line $13$: the test is included if its bin is to the left of a bin selected using the bin weights as selection probabilities.

The above paragraphs describe the training data sampler completely. The proposed ideas are certainly not the only possible ones for a reasonable sampler. In the first paper describing WOGAN \cite{sbst_long}, the proposed sampler was slightly different: it always had $Q = 1$, the bin weight computation involved a sigmoid function, and new tests were not treated differently. The problem with having $Q = 1$ is that if the robustness value range is significantly smaller than $[0, 1]$, all tests can end up in a single bin. This was not a problem in the experiment of \cite{sbst_long} as the robustness range was the full interval $[0,1]$. Anyway, the results of \autoref{ssec:rq2} show that the WOGAN algorithm is robust to minor changes in the sampler, so minor variations to the proposed ideas are likely to be inconsequential.

\begin{algorithm}[t]
  \caption{Online training of a WGAN.}\label{alg_sampler}
  \begin{algorithmic}[1]
    \Require{WGAN training epochs $E_W$, sample size $m$, number of bins $N_B$.}
    \Statex
    \Function{train\_wgan\_online}{$W$; $S$; $T$, NEW; $R$}
        \For{$i \in \{1, \ldots, E_W\}$}
            \State sample\_size $\gets \min\{|T|, m\}$
            \State $X \gets $ \Call{$S$}{$T$, NEW; R, sample\_size}
            \State \Call{train\_wgan}{$W$; $X$}
        \EndFor
    \EndFunction
    \Statex
    \Function{quantile\_sampler}{$T$, NEW; $R$, sample\_size}
        \State sample $\gets \emptyset$
        \State $Q \gets$ \Call{compute\_quantile}{$R$}
        \State bins $\gets$ \Call{bin\_tests}{$T$; $Q$}
        \State weights $\gets$ \Call{compute\_weights}{bins}
        \For{test $\in $ NEW}
            \State $i \gets $ \Call{find\_bin\_index}{test}
            \If{$i \leq $ \Call{sample\_bin\_index}{weights}}
                \State sample $\gets$ sample $\cup \, \{$test$\}$
                \State \Call{remove\_test\_from\_bin}{i, test}
            \EndIf
        \EndFor
        \For{$j \in \{1, \ldots, \, $sample\_size$ \, - \, |$sample$|\}$}
            \State $i \gets $ \Call{sample\_bin\_index}{weights}
            \While{$|$bins$[i]| = 0$}
                \State $i \gets i + 1$
            \EndWhile
            \State test $\gets$ \Call{select\_from\_bin}{i}
            \State \Call{remove\_test\_from\_bin}{i, test}
            \State sample $\gets$ sample $\cup \, \{$test$\}$
        \EndFor
        \State \Return sample
    \EndFunction
\end{algorithmic}
\end{algorithm}

We remark that the training data sampling implicitly assumes that $\rho$ takes a multitude of values. Indeed, if not, then it is possible that already for large $Q$ the training data consists of tests with constant robustness $r$. In such a case, it is intuitively likely that the generator ends up only producing tests with robustness $r$ and never improves. Also, if the majority of the tests have robustness $r$, then the analyzer likely fails to generalize, and a test with robustness less than $r$ is unlikely to be included in the training data even if the generator manages to generate such a test. 

\subsection{Sampling WOGAN Generators}\label{ssec:sampling_wogan_generators}
Executing the WOGAN RFG-algorithm yields a generator $\generator$ as in \autoref{ssec:rfg_problem}, and it needs to be sampled for tests. This can be done as in \autoref{ssec:rfg_problem} by sampling the latent space $\latentspace$ and mapping the samples via $\generator$. If the generator is trained very well, this suffices. In practice, however, we expect the generator to occasionally output high-robustness tests. Since an execution of the WOGAN algorithm also produces a trained analyzer $A$, rejection sampling can be used. We thus propose to use \autoref{alg_rejection_sampler} to sample WOGAN generators.

\section{Research Questions}\label{sec:research_questions}
We set out to evaluate the generators created by the WOGAN RFG-algorithm. In particular, we evaluate a practical WOGAN implementation described in \autoref{sec:wogan_setup}. The WOGAN generators are compared against the baseline uniform random generator, and they are sampled as described in \autoref{ssec:sampling_wogan_generators}. The evaluation is based on the evaluation metrics given in \autoref{sec:evaluation_methodology}. The metrics include a diversity metric and a quantile metric. The latter measures the capability of a generator to produce low-robustness tests, and it is used to rank generators from best to worst in this capability. Our first research question is as follows.

\begin{itemize}
  \item[] \textbf{RQ1}. What is the diversity and rank of the WOGAN generators compared to the baseline uniform random generator?
\end{itemize}

The WOGAN algorithm is complex and has several features to accomplish the training of the WGAN and the analyzer from scratch. The features include sampling a partially trained WGAN using the analyzer (\autoref{ssec:wgan_sampling}) and the WGAN training data sampler (\autoref{ssec:wgan_training_online}). These features were introduced to produce certain effects, and it is vital to empirically verify the presence of the effects. This leads to the following research question.

\begin{itemize}
  \item[] \textbf{RQ2}. What are the effects of the proposed WGAN training data sampler and the analyzer on the performance of the trained WOGAN generator? What is the effect of the rejection sampling on WOGAN generator samples?
\end{itemize}

The aim of an RFG-algorithm is to produce a $D_0$-generator. While this is not feasible in practice, it is nevertheless interesting to see how many falsifying tests a sample from a generator contains. In this context, it is natural to check how frequently a generator sample contains a falsifying test and compare the results to samples from uniform random sampling and existing RF-algorithms. Since the WOGAN generators are designed to be diverse, it is also interesting to see how diverse falsifying tests a generator can generate.

\begin{itemize}
  \item[] \textbf{RQ3}. How frequently do WOGAN generators produce falsifying tests, and how does this frequency compare to the state-of-the-art RF-algorithms? How diverse are the falsifying tests found by the WOGAN generators?
\end{itemize}

The WOGAN generator training and sampling is done with a fixed SUT test execution budget. It is also of interest to know what are the computational resources needed to run the WOGAN algorithm and to train the models. It is also important to know what is the time cost of using a WOGAN generator over a simple random generator. Thus we ask the following question.

\begin{itemize}
  \item[] \textbf{RQ4}. What is the computational cost of the WOGAN algorithm and the computational overhead of using a WOGAN generator in terms of used time?
\end{itemize}

\section{Evaluation Methodology}\label{sec:evaluation_methodology}
In this section, we describe how we evaluate and compare generators. Ideally we would like to
state principles for judging how close a generator is being a $D_0$-generator. However, practical principles would require us either to have a $D_0$-generator or to understand the set $\critical_\varepsilon$ for some small $\varepsilon$. The SUTs we consider are so complex that this is infeasible. Therefore we propose metrics that focus on two key properties of $D_0$-generators and state principles according to which we can decide if a generator sample is closer to a sample from a $D_0$-generator than a sample from another generator. The properties are: i) generation of tests with low robustness (\autoref{ssec:quantile_scores}) and ii) input diversity (\autoref{ssec:diversity}). In \autoref{ssec:ranking}, we explain how we rank generators. \autoref{ssec:falsification_capability} describes how we evaluate a generator sample in terms of falsifying tests it contains.

Throughout this section, let $\sut\colon \inputspace \to \outputspace$ be a SUT with an associated robustness metric $\rho$ as in \autoref{sec:problem_description}. We let $T$ denote a sample from a generator $\generator$, that is, $T$ is a collection of pairs $(t, \rho(t))$ where $t \in \inputspace$ and $\rho(t) = \rho(\sut(t))$.

\subsection{Quantile Scores}\label{ssec:quantile_scores}
It is unreasonable to evaluate a sample $T$ by computing $\max_{t \in T} \rho(t)$ as
this number can be large even when $\rho(t)$ is typically low. To better capture the nature of the distribution
of $\rho \circ \sut \circ \generator$ in numbers, we propose to measure $T$ based on two \emph{quantile scores} $Q_L$
and $Q_U$ as follows. Let $q_L \in [0, \tfrac12]$ and $q_U = 1 - q_L$. We set
\begin{equation*}
  Q_L = \min_z |\{t \in T : \rho(t) \leq z\}| \geq q_L |T| \quad \text{and} \quad Q_U = \min_z |\{t \in T : \rho(t) \leq z\}| \geq q_U |T|.
\end{equation*}
These numbers estimate the $q_L$ and $q_U$ quantiles of the distribution of $\rho \circ \sut \circ \generator$, that
is, the numbers $z_L$ and $z_U$ such that
\begin{equation*}
  \prob{\rho \circ \sut \circ \generator \leq z_L} = q_L \quad \text{and} \quad \prob{\rho \circ \sut \circ \generator \leq z_U} = q_U.
\end{equation*}
If $\generator$ is a $D_\varepsilon$-generator, then $Q_L = q_L\varepsilon$ and $Q_U = q_U\varepsilon$. This yields the heuristic that among two generators the better one has smaller $Q_L$ and $Q_U$. To better understand the shape of the distribution, we report both $Q_L$ and $Q_U$. In the evaluation of \autoref{sec:results}, we select $q_L = 0.25$.

\subsection{Diversity}\label{ssec:diversity}
The aim of this section is to define a diversity measure for a sample $T$. Recall that the diversity, as discussed in \autoref{ssec:rfg_problem}, concerns only the input space $\inputspace$ of the SUT, so we do not need to consider the system outputs.

\subsubsection{Similarity Measure}
A \emph{similarity measure} $\similarity$ is a symmetric function $\inputspace \times \inputspace \to [0,1]$ which
tells to what degree two tests are similar. Value $1$ represents maximally similar tests
(including equality) and $0$ represents maximal dissimilarity. Let $B$ in $[0,1]$ be a fixed \emph{similarity threshold}.
We say that two tests $t_1$ and $t_2$ are \emph{similar} if $\similarity(t_1, t_2) \geq B$. Otherwise we say that the
tests are \emph{dissimilar}. The similarity measures used in this paper are described in \autoref{ssec:similarity_measures}. In \autoref{sec:results}, we use $B = 0.9$.

\subsubsection{Sample Clustering}
Next we propose how to cluster the tests in $T$ according to a similarity measure $\similarity$. In order to make the
clustering deterministic and noninformative (that is, unrelated to the similarity measure or the order of tests in
$T$), we order the tests of $T$ as $t_1 \leq t_2 \leq \dotsm \leq t_n$, $n = |T|$, according to a deterministic
order relation $\leq$. In this paper, we use the lexicographic order. Say two tests $t_1$ and
$t_2$ are respectively determined by the vectors $(x_1, \ldots, x_D)$ and $(y_1, \ldots, y_D)$.
Then $t_1 < t_2$ if there exists $i$ such that $x_j = y_j$ for all $j$ such that $j < i$
and $x_i < y_i$ (here $<$ is the natural order on the reals). We then construct the clustering greedily according to
\autoref{alg2}. The obtained clustering has the property that $\similarity(t_1, t_2) \geq B$ for
any two tests $t_1$ and $t_2$ belonging to the same cluster. The proposed clustering is a form of hierarchical clustering
\cite[Sec.~14.3.12]{eosl}. See \autoref{ssec:similarity_measures} for visualizations of clusters.

\begin{algorithm}[t]
  \caption{Greedy sample clustering algorithm.}\label{alg2}
  \begin{algorithmic}[1]
    \Require{Ordered tests $t_1 \leq t_2 \leq \dotsm \leq t_n$, similarity measure $\similarity$, similarity threshold $B$.}
    \Statex
    \State $\textup{clusters} \gets \{\}$
    \For{$i \in \{1, \ldots, n\}$}
      \State $\textup{found} \gets \textup{false}$
      \For{$\textup{cluster} \in \textup{clusters}$}
        \State $\textup{found} \gets \textup{true}$
        \For{$t \in \textup{cluster}$}
          \State $\textup{similarity} \gets \similarity(t_i, t)$
          \If{\textup{similarity} < B}
            \State $\textup{found} \gets \textup{false}$
            \State $\textup{break}$
          \EndIf
        \EndFor
        \If{\textup{found}}
          \State $\textup{cluster} \gets \textup{cluster} \cup \{t_i\}$
          \State $\textup{break}$
        \EndIf
      \EndFor
      \If{\textup{not found}}
        \State $\textup{clusters} \gets \textup{clusters} \cup \{\{t_i\}\}$
      \EndIf
    \EndFor
\end{algorithmic}
\end{algorithm}

\subsubsection{Diversity Score and Diversity Loss}\label{ssec:diversity_score_loss}
Given a sample $T$ and its clustering $\{C_1, \ldots, C_k\}$ as described in the preceding subsection, we define the
\emph{diversity score} $\diversity(T)$ of $T$ to be $k/|T|$ (if $T = \emptyset$, we set $\diversity(T) = 0$). The higher $\diversity(T)$
is, the higher diversity we consider $T$ to have. Notice however that the diversity score is relative to the sample
size and typically tends to $0$ as $|T| \to \infty$. Clearly if $T$ is generated by a collapsed generator, then
$\diversity(T)$ is low. In order to determine how well $T$ represents a sample from a $D_\varepsilon$-generator, we
propose to compare $\diversity(T)$ to $\diversity(R)$ where $R$ is a sample of size $|T|$ obtained by sampling $\inputspace$
uniformly randomly. We cannot know what is the diversity score of a $D_\varepsilon$-generator (for a sample of size
$|T|$), but $\diversity(R)$ represents an upper bound for it. We propose to rate the \emph{diversity loss} $\divloss(T)$ of $T$
compared to a $D_\varepsilon$-generator by setting $\divloss(T) = \diversity(T)/\diversity(R)$. We define the diversity
loss to be \emph{negligible}, \emph{small}, \emph{moderate}, and $\emph{large}$ if respectively
$\divloss(T) \in [0.98, 1.00]$, $\divloss(T) \in [0.75, 0.98)$, $\divloss(T) \in [0.50, 0.75)$, and
$\divloss(T) \in [0.00, 0.50)$.

\subsection{Generator Ranking}\label{ssec:ranking}
We rank two generators $\generator_1$ and $\generator_2$ based on the quantile scores computed from generator samples as follows. For each generator, we generate $N$ independent samples of size $M$. We compute the quantile scores for these samples as described in \autoref{ssec:quantile_scores}. We average the quantile scores to obtain the means
$\overline{Q}_L(\generator_1)$ and $\overline{Q}_U(\generator_1)$ with standard deviations $S_L(\generator_1)$ and $S_U(\generator_1)$ for
$\generator_1$ and similarly for $\generator_2$. Let also
$D_L = \smash[t]{\overline{Q}}_L(\generator_2) - \smash[t]{\overline{Q}}_L(\generator_1)$ and
$D_U = \smash[t]{\overline{Q}}_U(\generator_1) - \smash[t]{\overline{Q}}_U(\generator_2)$.

We write $L(\generator_1) \prec L(\generator_2)$ if
$\smash[t]{\overline{Q}_L(\generator_1)} + S_L(\generator_1) < \smash[t]{\overline{Q}_L(\generator_2) - S_L(\generator_2)}$. If neither
$L(\generator_1) \prec L(\generator_2)$ nor $L(\generator_2) \prec L(\generator_1)$ holds, then we write $L(\generator_1) \sim L(\generator_2)$.
Similarly we define the relation $U(\generator_1) \prec U(\generator_2)$ with the condition
$\smash[t]{\overline{Q}}_U(\generator_1) + S_U(\generator_1) < \smash[t]{\overline{Q}}_U(\generator_2)- S_U(\generator_2)$. Observe that
$L(\generator_1) \sim L(\generator_2)$ when the intervals defined by the endpoints $\smash[t]{Q}_L(\generator_1) \pm S_L(\generator_1)$ and
$\smash[t]{Q}_L(\generator_2) \pm S_L(\generator_2)$ intersect. Analogous holds when $U(\generator_1) \sim U(\generator_2)$.

We define the relation
$\generator_1 \prec \generator_2$ by the following rules.
\begin{enumerate}[(i)]
  \item If $L(\generator_1) \prec L(\generator_2)$ and $U(\generator_1) \prec U(\generator_2)$, then $\generator_1 \prec \generator_2$.
  \item If $L(\generator_1) \sim L(\generator_2)$ and $U(\generator_1) \prec U(\generator_2)$, then $\generator_1 \prec \generator_2$.
  \item If $L(\generator_1) \prec L(\generator_2)$ and $U(\generator_1) \sim U(\generator_2)$, then $\generator_1 \prec \generator_2$.
  \item If $L(\generator_1) \prec L(\generator_2)$ and $U(\generator_1) \succ U(\generator_2)$ and $2D_L > D_U$, then $\generator_1 \prec \generator_2$.
  \item If $L(\generator_1) \prec L(\generator_2)$ and $U(\generator_1) \succ U(\generator_2)$ and $2D_L < D_U$, then $\generator_1 \succ \generator_2$.
\end{enumerate}
If neither $\generator_1 \prec \generator_2$ nor $\generator_2 \prec \generator_1$ holds, then we write $\generator_1 \sim \generator_2$. Notice that in
the conditions (iv) and (v), we value improvement in $Q_L$ twice as much as improvement in $Q_U$.

Let us now rank generators using the above relation $\prec$ (which is not a transitive relation). We do this using a
form of tournament ranking. Given generators $\generator_1$, $\ldots$, $\generator_n$, we define the \emph{count} $C(\generator_i)$ as
the number $|\{j : \text{$\generator_i \prec \generator_j$ or $\generator_i \sim \generator_j$}\}|$. Then we assign rank $1$ to all $k_1$
generators with the highest count. The generators with the second highest count get rank $1 + k_1$ and so on. For
example, consider the line of \autoref{tbl:generator_quality} for the $\AFC$ benchmark (the $Q$-scores). Then
the counts for the generators (from left to right) are $4$, $3$, $4$, $4$, so the generators have ranks $1$,
$2$, $1$, $1$.

\subsection{Falsification Rate and Diversity}\label{ssec:falsification_capability}
Let us now evaluate a generator $\generator$ in terms of falsifying tests it can generate. Let $T_1$, $\ldots$, $T_N$ be independent samples of fixed size $B$ obtained by sampling $\generator$ (or, more generally, by executing an RF-algorithm several times). The \emph{falsification rate} is defined as the number $K/N$, where $K$ is the number of samples containing a falsifying test. A generator $\generator_1$ achieving a higher falsification rate than a generator $\generator_2$ is more \emph{effective}: $\generator_1$ falsifies more frequently than $\generator_2$.

Let $F_i$ be the set of falsifying tests for the sample $T_i$. We set the \emph{falsification diversity} $D_F(\generator)$ of $\generator$ to be the mean of the numbers $\diversity(F_i)$, $i = 1, \ldots, N$. We do not consider diversity loss in the context of falsification.

\section{Benchmark Selection}\label{sec:benchmark_selection}
To answer the research questions, we conduct several experiments where we compare the WOGAN generators against baseline generators or generators created by variants of the WOGAN algorithm itself.
The comparisons are done on several benchmarks that represent a varied selection
of cyber-physical systems. These include selected benchmarks from the ARCH-COMP 2023 competition \cite{ARCH23} and two
benchmarks that test the LKAS of a autonomous car derived from the SBFT 2023 CPS tool competition
\cite{SBFT-toolcomp23}.

\subsection{ARCH-COMP 2023 Benchmarks}
ARCH-COMP 2023 \cite{ARCH23} is a friendly CPS falsification competition where the goal is to submit an RF-algorithm
for the falsification of certain preselected requirements. The falsification competition has
been held since 2018; see \cite{ARCH21,ARCH22} for the reports of the latest competitions. The ARCH23 benchmarks have
been used for the evaluation of requirement falsification algorithms on numerous occasions; see
\cite{ARCH21,ARCH22,ARCH23,2021:efficient_optimization_based_falsification_of_cyber,2021:falsification_of_hybrid_systems_using_adaptive,2021:part_x_a_family_of_stochastic_algorithms,2023:search_based_software_testing_driven}
to name a few.

We have selected the benchmarks $\AFC$, $\ATI$, $\ATVIA$, $\ATVIB$, $\ATVIC$, $\CCIII$, $\CCIV$,
$\FA$, $\NN_{0.03}$, $\NN_{0.04}$ from \cite{ARCH23}. We have excluded benchmarks that are easy to falsify: we
excluded a benchmark of \cite{ARCH23} if uniform random search could falsify it with a falsification rate greater than
$0.5$ with $75$ executions over $50$ repetitions. We have also excluded the benchmarks $\textup{AFC33}$ and
$\textup{SC}$ because at most one requirement falsification algorithm in \cite{ARCH23} could falsify them. Finally, we exclude the benchmark $\textup{PM}$ because the robustness metric is discrete taking only few values; cf. the last paragraph of \autoref{ssec:wgan_training_online}.

Below we describe the $\textup{AT}$ and $\FA$ benchmarks and their input formats (the text is almost verbatim from \cite{ogan_full}). The remaining benchmarks and their input formats are described in the competition report \cite{ARCH23}; see also \cite{ogan_full}. The requirements for the benchmarks are listed in \autoref{tbl:req}.

\begingroup
\setlength{\tabcolsep}{6pt} 
\renewcommand{\arraystretch}{1.2} 

\begin{table}
\begin{center}
\begin{tabular}{ll}
  \toprule
  Req. & Formula \\
  \midrule
  \midrule
  $\textup{AFC27}$ & $\always_{t \in [11,50]} \left( (\mathrm{rise}(t) \lor \mathrm{fall}(t)) \rightarrow (\always_{t' \in [1,5]}  |\signal[t']{\mu}| < \beta) \right)$ \\
                   & $\mathrm{rise}(t) = (\signal[t]{\theta} < 8.8) \land (\eventually_{t' \in [0,0.05]} \signal[t']{\theta} > 40.0)$ \\
                   & $\mathrm{fall}(t) = (\signal[t]{\theta} > 40.0) \land (\eventually_{t' \in [0,0.05]} \signal[t']{\theta} < 8.8)$ \\
                   & $\beta = 0.008$ \\
  \midrule
  $\textup{AT1}$              & $\always_{t \in [0,30]} \, \signal[t]{v} < 120$ \\
  $\textup{AT6}_{T,v,\omega}$ & $(\always_{t \in [0,30]} \signal[t]{\omega} < \omega) \rightarrow (\always_{t \in [0,T]} \signal[t]{v} < v)$ \\
  \midrule
  $\textup{CC3}$ & $\always_{t \in [0,80]} \left( ( \always_{t' \in [0,20]} \, \signal[t']{Y_2} - \signal[t']{Y_1} \leq 20 ) \lor (\eventually_{t' \in [0,20]} \signal[t]{Y_5} - \signal[t]{Y_4} \geq 40) \right)$ \\
  $\textup{CC4}$ & $\always_{t \in [0,65]} \eventually_{t' \in [0,30]} \always_{t'' \in [0,20]} \, \signal[t'']{Y_5} - \signal[t'']{Y_4} \geq 8$ \\
  \midrule
  $\textup{F16}$ & $\always_{t \in [0,15]} \, \textup{ALTITUDE}(t) > 0$ \\
  \midrule
  $\textup{NN}_\beta$ & $\always_{t \in [1,37]} (|\signal[t]{P} - \signal[t]{R}| > \alpha + \beta|\signal[t]{R}|$ \\
                      & $\rightarrow \eventually_{t' \in [0,2]} \always_{t'' \in [0,1]} |\signal[t'']{P} - \signal[t'']{R}| < \alpha + \beta|\signal[t'']{R}|)$ \\
                      & $\alpha = 0.005$ \\
  \midrule
  $\textup{AMBIEGEN}_d$ & $\always_{t} \, D(t) \leq 4.40$ \\
                        & \text{$d$ is the number of input curvature values} \\
  \midrule
  $\textup{BEAMNG}_d$ & $\always_{t} \, \mathrm{BOLP}(t) \leq 0.95$ \\
                      & \text{$d$ is the number of input curvature values}
\end{tabular}
\caption{STL requirements for the selected benchmarks.}\label{tbl:req}
\end{center}
\end{table}

\endgroup

\textbf{Automatic Transmission (AT)} \cite{2015:benchmarks_for_temporal_logic_requirements_for_automotive}.
Given two input signals $\textup{THROTTLE}$ and
$\textup{BRAKE}$, the model outputs three signals: $v$ (speed of the modeled car), $\omega$ (engine speed of the car in RPM), and $g$ (currently selected gear). As in \cite{ARCH23}, we assume that $\textup{THROTTLE}(t) \in [0, 100]$ and $\textup{BRAKE}(t) \in [0, 325]$ for all $t$ during the simulation time of $30$ time units. We assume that the input signals are piecewise constant signals consisting of $6$ segments of $5$ time units as in \cite{ARCH23}. For the output signals, we set the following ranges: $\signal[t]{v} \in [0, 125]$, $\signal[t]{\omega} \in [0, 4800]$, and $\signal[t]{g} \in \{1,2,3,4\}$ for all $t$. The requirements $\ATI$ and $\textup{AT6}_{T,v,\omega}$ are found in \autoref{tbl:req}.

\textbf{F-16 Ground Collision Avoidance (F16)} \cite{2018:verification_challenges_in_f16_ground_collision_avoidance}. The model controls an aircraft and attempts to avoid ground collisions. The input for the model is a vector with three components $\textup{ROLL}$, $\textup{PITCH}$, and $\textup{YAW}$ which determine the orientation of the aircraft in the starting position at the altitude of $4040$ feet. As in \cite{ARCH23}, we set $\textup{ROLL} \in [0.2\pi, 0.2833\pi]$, $\textup{PITCH} \in [-0.4\pi, -0.35\pi]$, and $\textup{YAW} \in [-0.375\pi, -0.125\pi]$. The output of the model is a signal $\textup{ALTITUDE}$ taking values in $[0,4040]$ and representing the altitude of the aircraft during the simulation of $15$ time units. The requirement $\FA$ is listed in \autoref{tbl:req}. We use the Matlab implementation of the model as in \cite{ARCH23}.

\subsection{The BeamNG.tech Benchmark}\label{ssec:beamng_tech}
\begin{figure}
  \begin{minipage}{0.49\columnwidth}
    \begin{center}
    \includegraphics[width=\linewidth]{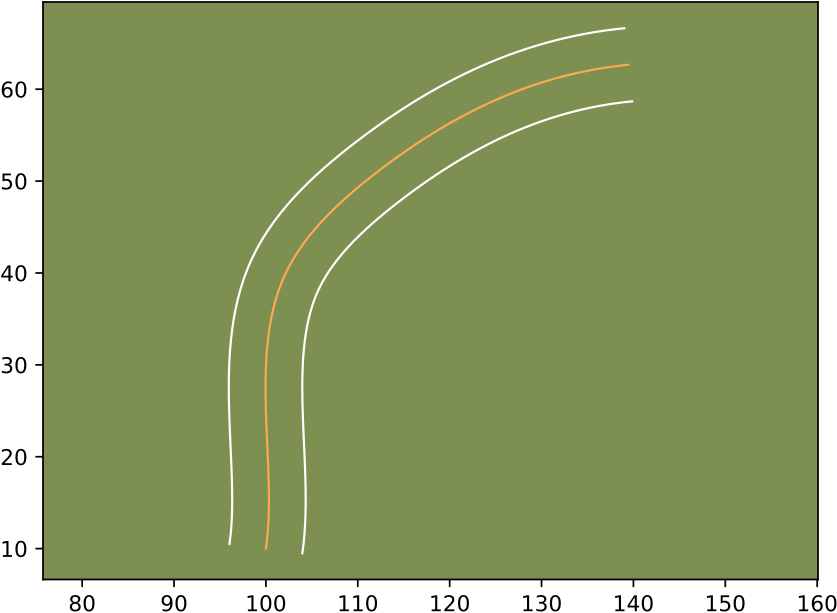}
    \end{center}
  \end{minipage}%
  \begin{minipage}{0.49\columnwidth}
    \begin{center}
    \includegraphics[width=0.9\linewidth]{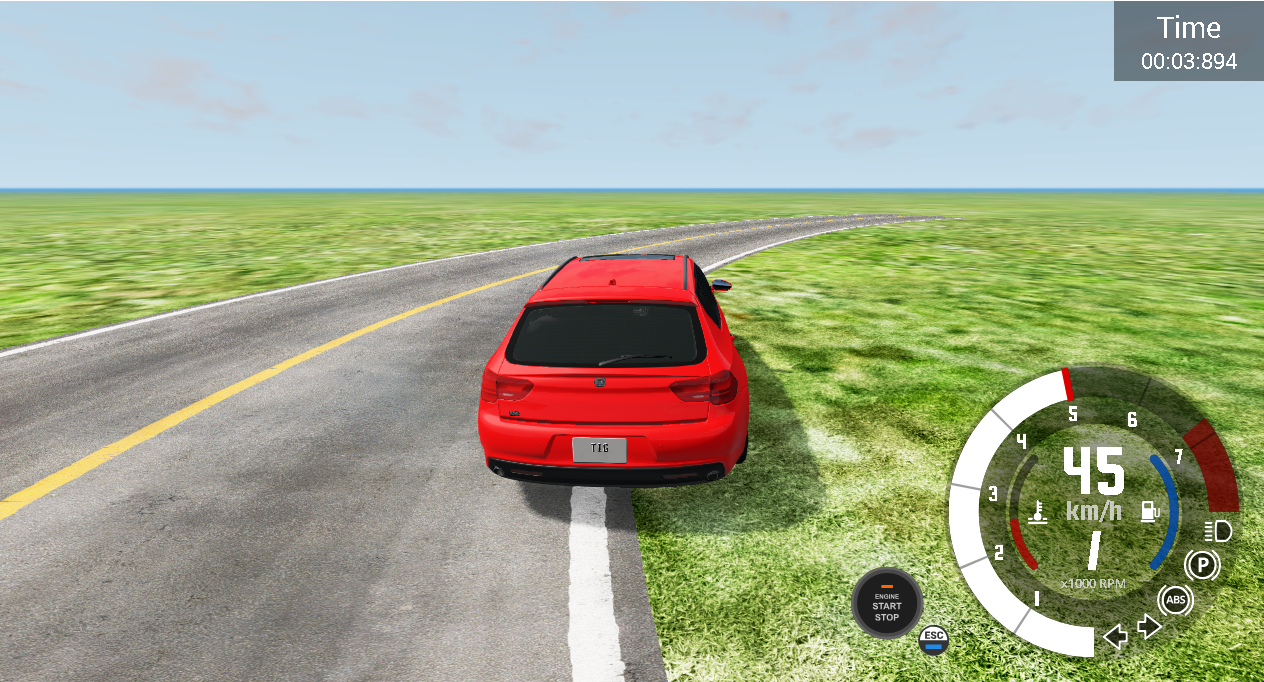}
    \end{center}
  \end{minipage}%
  \caption{\textmd{A virtual road and the BeamNG.tech agent driving on it.}}\label{fig:simulator}
\end{figure}

The goal of the SBFT 2023 CPS tool competition \cite{SBFT-toolcomp23} was to test the lane keeping assist system (LKAS) of an autonomous car in a simulated environment. See \cite{SBST-toolcomp21,SBST-toolcomp22,SBFT-toolcomp23,SBFT-toolcomp24} for recent installments of the competition. The simulator is the BeamNG.tech\footnote{\url{https://beamng.tech}} simulator version 0.26.2.0 developed by the BeamNG company. The LKAS is tested by providing virtual roads which the driving agent is supposed to drive, from beginning to end, while keeping the right lane. The virtual roads are defined by control points and interpolated into complete roads by the simulator. The driving environment of the agent is plain, containing only a two-lane road on a flat grass field. See \autoref{fig:simulator} for a virtual road and to see how it appears in the simulator. The simulation output is a body out of lane percentage (BOLP) signal which measures the proportion of the car outside its designated lane at any given time. The requirement to be falsified, as in \cite{SBFT-toolcomp23}, is $\always_{t} \, \mathrm{BOLP}(t) \leq 0.95$, which states that at most $95 \%$ of the car can be outside the lane at any time. A requirement involving also the driving angle of the agent is considered in \cite{2024:testing_cyber_physical_systems_with_explicit}, but we do not consider it here. We remark that the simulation is rather slow: evaluating a single test typically takes $30$ to $40$ seconds.

The virtual roads are be defined by a variable number of control points which need to fit into a map of $200 \times 200$ units. Since WOGAN expects fixed-dimensional input, we represent a road as a vector $(c_1, \ldots, c_d)$ in $[-0.07, 0.07]^d$ (we use the values $d \in \{5, 7, 9\}$). We have chosen a representation based on curvature values \cite{frenetic}. In short, we assume that each road begins at the initial control point $(100, 10)$ and initially faces up. The number $c_1$ describes the curvature (the rate of turning) of the road when $15$ units are traveled from the initial location in unit time at unit speed. This determines the second control point. The procedure is repeated for the remaining curvature values $c_2$, $\ldots$, $c_d$ so that a total of $d+1$ control points are determined. These resulting control points are then fed as an actual input to the simulator.

Additionally a virtual road must be valid which means that it must i) not self-intersect and ii) not contain overly sharp turns. The validity check is implemented by the SBFT 2023 CPS competition pipeline\footnote{Available at \url{https://github.com/sbft-cps-tool-competition/cps-tool-competition}.}. This means that a sample from a generator is possibly invalid. We have simply chosen to sample until the sampled test is valid. It follows that WOGAN needs to learn to generate not only low-robustness tests but also valid tests. The chosen input representation facilitates this task: sampling with the curvature representation is much more likely to yield a valid road than simply sampling plane points uniformly randomly \cite{sbst_long}.

The chosen input representation allows only to represent a subset of all possible roads: the more curvature values are allowed, the more complicated roads can be expressed. In the 2022 and 2023 editions of the SBFT competition, WOGAN successfully used the curvature input representation with $d = 5$. Here we also consider the values $d = 7$ and $d = 9$ to see how this affects the results.

\subsection{The AmbieGen Benchmark}
The AmbieGen tool \cite{2022:ambiegen_tool_at_the_sbst_2022_tool_competition,2022:a_search_based_framework_for_automatic_generation,2023:ambiegen_a_search_based_framework_for_autonomous} contains a simplified and fast-to-execute driving agent that was used as a surrogate model for the BeamNG.tech simulator driving agent in \cite{2022:ambiegen_tool_at_the_sbst_2022_tool_competition}. In the surrogate model, a road is again determined by control points, but the road is one-dimensional (no lanes) and the car is modeled as a point. The output of the model is a distance (D) of the car to the road during the simulation. Similar to the BeamNG.tech benchmark, we require that the car does not drive too far away from the road. The requirement we use is $\always_{t} \, \mathrm{D}(t) \leq 4.40$. The number $4.40$ is chosen to make the falsification difficult for uniform random search: uniform random search achieves a falsification rate less than $0.20$ with $300$ SUT evaluations over $50$ replications.

We use the same curvature-based input representation as in the BeamNG.tech benchmark. Again each road must be valid for it to be simulated; a valid road must i) not self-intersect and ii) not contain too sharp turns. The validity check is implemented in the AmbieGen tool. The code is available at \url{https://github.com/dgumenyuk/tool-competition-av}.

\subsection{Input Representation and Similarity Measures}\label{ssec:similarity_measures}
Here we describe the similarity measures selected for the benchmarks. The similarity measures ultimately determine in what sense do we consider inputs to be diverse. Before the description, we need to discuss how we represent system inputs.

As already present in the above discussion on ARCH-COMP 2023 benchmarks, we represent signal inputs as piecewise constant signals of equal piece length. Such a signal consisting of $D$ pieces and taking values in an interval $[A, B]$ is fully described by a vector of length $D$. We map such a vector to a vector with components normalized to the interval $[-1, 1]$ via the linear map $x \mapsto (-2x + A + B)/(A-B)$. We thus take the input space $\inputspace$ to be $[-1, 1]^D$. Given an input in $\inputspace$, we construct the input signal by first applying the inverse transformation and then by sampling the piecewise signal according to a predefined sampling frequency, which is fixed to $0.01$ time units in this article. If the system input consists of several signals, we simply concatenate the vectors representing each signal. If the input is not a signal but a vector, as in the F16, BeamNG.tech, and AmbieGen benchmarks, we do the same as above, but we omit the sampling step.

For all benchmarks, we set the latent space dimension $d_H$ to $10$. The dimensions of the benchmark input spaces take values in $\{3, 5, 7, 9, 11, 12, 40\}$. It is easier to learn a mapping from a higher-dimensional space to a lower-dimensional space, so we compromise and set $d_H = 10$.

\subsubsection{Similarity Measure for the ARCH-COMP 2023 Benchmarks}
\begin{figure}
  \begin{minipage}{0.49\columnwidth}
    \begin{center}
    \includegraphics[width=\linewidth]{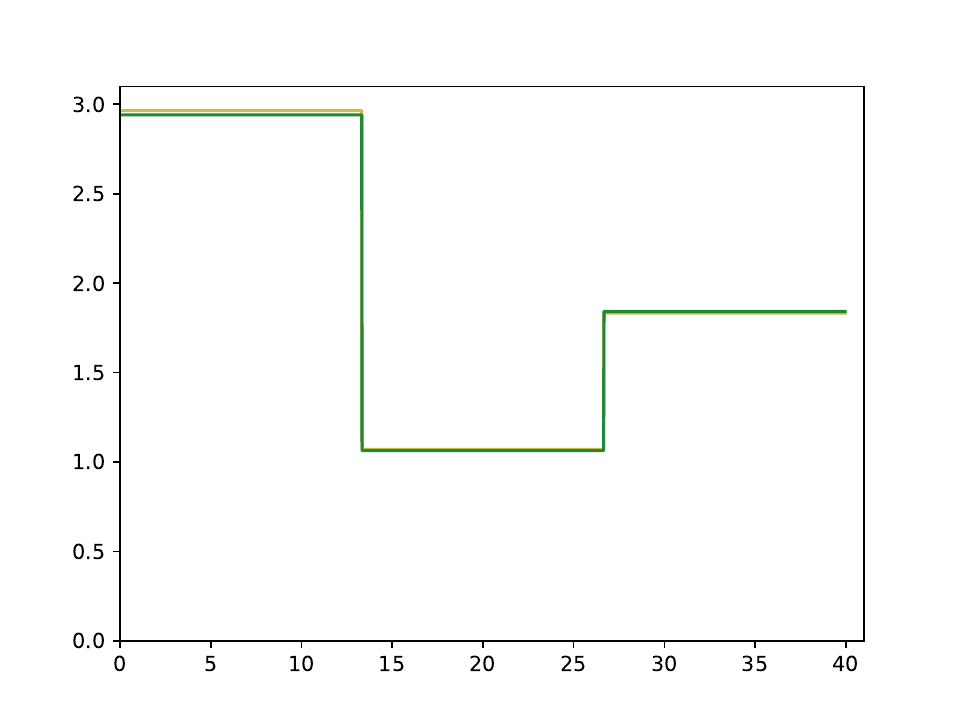}
    \end{center}
  \end{minipage}%
  \begin{minipage}{0.49\columnwidth}
    \begin{center}
    \includegraphics[width=\linewidth]{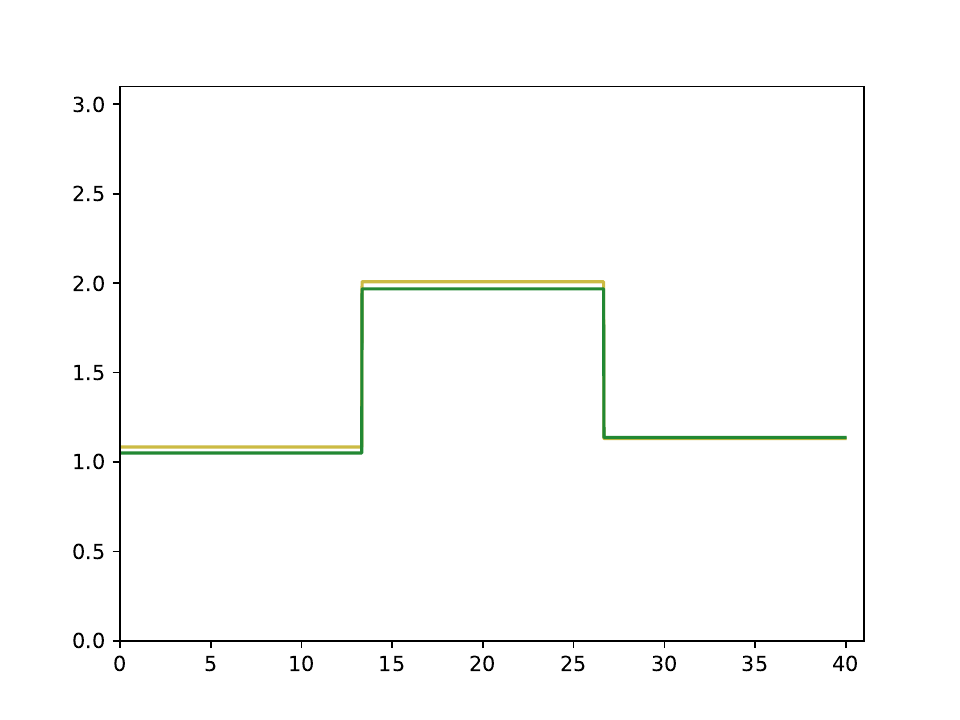}
    \end{center}
  \end{minipage}%
  \caption{Four input signals that form two clusters for the $\textup{NN}_{0.03}$ benchmark.}
  \label{fig:cluster_arch23}
\end{figure}

For two tests $t_1$ and $t_2$, represented as vectors $(x_1, \ldots, x_D)$ and $(y_1, \ldots, y_D)$, we define their
similarity measure $\similarity$ by setting
\begin{equation*}
  \similarity(t_1, t_2) = 1 - 2 \max_i |x_i - y_i|.
\end{equation*}
When the test represents a piecewise signal, the tests are considered similar with similarity bound $B = 0.95$ if all signal pieces differ at most $5 \%$ from each other. See \autoref{fig:cluster_arch23} for four input signals from two clusters formed with the described similarity measure.

\subsubsection{Similarity Measure for the BeamNG.tech and AmbieGen Benchmarks}
\begin{figure}
  \includegraphics[width=\linewidth]{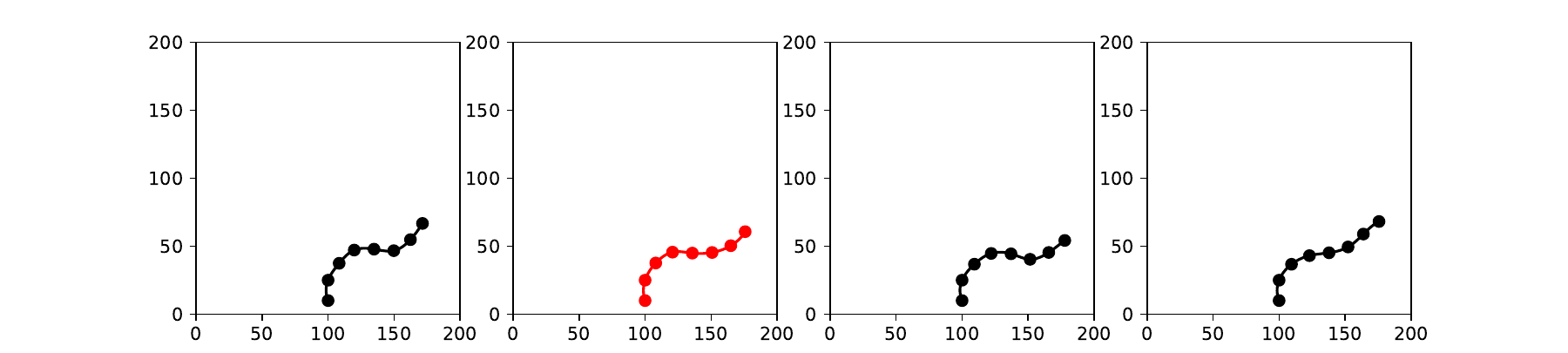}
  \includegraphics[width=\linewidth]{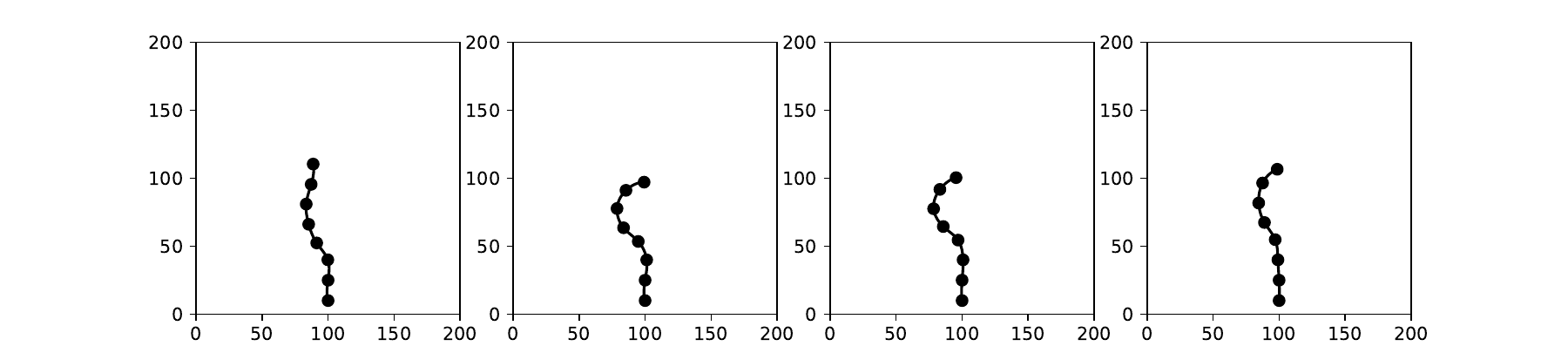}
  \caption{Tests from two clusters for the BeamNG benchmark when $d = 7$; one cluster per row. Red color indicates a falsifying input; cf. \autoref{tbl:req}.}
  \label{fig:cluster_beamng}
\end{figure}

A test for the BeamNG.tech and AmbieGen benchmarks is a vector $(c_1, \ldots, c_d)$ of curvature values that are transformed into a sequence $(p_1(t), \ldots, p_{d+1}(t))$ of control points that define a road. We define the distance $d(t_1, t_2)$ between tests $t_1$ and $t_2$ as
\begin{equation*}
  \sum_{i=1}^{d+1} \lVert p_i(t_1) - p_i(t_2) \rVert
\end{equation*}
where $\lVert \cdot \rVert$ is the Euclidean norm. We define a similarity measure $\similarity$ by setting
\begin{equation*}
  \similarity(t_1, t_2) = 1 - \frac{d(t_1, t_2)}{K_d(d+1)}
\end{equation*}
where $K_d$ ensures that $\similarity$ has range contained in $[0,1]$. When $d = 5$, we select $K_d = 39.79/(d+1)$, which is slightly above the largest distance to the test $(-1, -1, -1, -1, -1)$ observed during a grid search over the input space. For $d = 7$ and $d = 9$, we respectively select $K_d = 51.57/(d+1)$ and $K_d = 66.89/(d+1)$. See \autoref{fig:cluster_beamng} for roads that belong to two different clusters.

\section{Machine Learning Models}\label{sec:wogan_setup}
Here we explain what sort of structure we have chosen to use for the machine learning models used by the WOGAN algorithm. We explain their training process and describe a setup that has good overall performance. We have not tuned the hyperparameters separately to each benchmark.

\subsection{The WGAN Model}
In \autoref{alg_wgan}, we use $\lambda = 10$ and $n_{\textup{critic}} = 5$ as recommended in \cite{wgan_gp}. For the online training of WGAN (\autoref{alg1}), we set the batch size $m$ to $32$ and we train for $2$ epochs ($E_W = 2$), keeping this number low as the training data is small.

The generator and critic models are neural networks; appropriate sigmoidal activation functions are used to ensure that the outputs map to the correct space. Since the execution budgets are low, the training data is rather small, so we opt to use small neural networks. The generator network is a a fully connected network with two hidden layers of $128$ neurons. The activation function for hidden layers is leaky ReLU with negative slope $0.01$ and batch normalization is used in every layer. The critic model has the same structure as the generator. 

We use the Adam optimizer for training both models. We use learning rate $0.0001$ and the default values $0.9$ and $0.999$ for the beta parameters. The loss functions are explained in \autoref{sec:wgan}.

\subsection{The Analyzer Model}\label{ssec:analyzer_model}
In all benchmarks with input signals, we use a convolutional network for the analyzer. We use two convolutional layers of $16$ feature maps with kernel size $2$, stride $1$, and $0$-padding $1$ followed by a leaky ReLU activation with negative slope $0.01$ and a maxpool layer with window size $2$ and stride $2$. The convolutional layer is flattened and mapped to a hidden layer of $128$ neurons (no activation). The output of this final layer is fed through the sigmoid function to produce a single number. Batch normalization is not used. The optimizer settings are as above. In the benchmarks where the input is a vector, we use the same network structure as above for the generator and the critic except that we use sigmoid as the output activation and we do not use batch normalization. The analyzer is trained for $10$ epochs on each execution of the main loop of \autoref{alg1}.

Since squared error does not penalize correctly when the estimated values are in the interval $[0, 1]$, we use the modified loss $\mathcal{L}$ of \cite{ogan_full}. We have
\begin{equation*}
  \mathcal{L}(\hat{y}, y) = (F(\hat{y}) - F(y))^2 + \lambda\left(F\left( \frac12 - \frac{\hat{y} - y}{2}\right) - F\left(\frac12\right) \right)^2
\end{equation*}
where $F(x) = \mathrm{logit}(0.98x + 0.01)$, $\mathrm{logit(x)} = \log x/(1-x)$, and $\lambda = 0.001$.

\section{Experiment Results}\label{sec:results}
In this section, we describe the experiments we have conducted to answer the research questions, and we analyze the results of the experiments.

\subsection{Data Collection and Experiment Replication}
The experiments related to the BeamNG.tech benchmark were performed on a machine running Windows 11 Professional with an Intel i9-12900 CPU, an NVIDIA GeForce RTX 3090 GPU, and 64 GB of RAM. The remaining experiments were conducted on a machine running Ubuntu 22.04 with an Intel i9-10900X CPU, an NVIDIA GeForce RTX 3090 GPU, and 64 GB of RAM.

The practical implementation of the WOGAN algorithm is part of the STGEM falsification framework, which is freely available at \url{https://gitlab.abo.fi/stc/stgem}. STGEM is written in Python, and we use Pytorch\footnote{\url{https://pytorch.org/}} to train the machine learning models. The benchmarks are available at \url{https://gitlab.abo.fi/stc/experiments/wogan} along with the experiment repeatibility instructions. The results necessary for recreating the tables and figures are available at \url{https://doi.org/10.5281/zenodo.13860526}.

\subsection{RQ1: Generator Evaluation}\label{ssec:rq1}
Recall that RQ1 asks to evaluate the generators created by WOGAN to see if they are diverse and rank better than the baseline uniform random generator, described below, on the falsification benchmarks from \autoref{sec:benchmark_selection}. The evaluation is based on the metrics proposed in \autoref{sec:evaluation_methodology}. We remark that the results for the $\BEAMNG_*$ benchmarks from the SBFT 2023 competition \cite{SBFT-toolcomp23} and the previous installments \cite{SBST-toolcomp22,SBST-toolcomp21} are not comparable to the results of WOGAN generators presented here. This is mainly because the goal of the competition is not to solve the RFG-problem and the used diversity metrics are different. In addition, a time budget and varying input representations were used, which prevent a direct comparison.\\

\noindent
\textbf{RANDOM} (random generator). This generator samples the SUT input space uniformly randomly. If a test needs to be valid to be executed on the SUT, resampling is done until a valid test is obtained. The diversity score of a sample of this generator indicates the maximum achievable diversity score for a generator sample (see \autoref{ssec:diversity_score_loss}).\\

The required WOGAN generators and analyzers were obtained by running the WOGAN algorithm on the selected benchmarks. We ran the WOGAN algorithm $N$ times with different initial conditions (different seeds for the random number generator), which resulted in $N$ WOGAN ge\-nera\-tors paired with $N$ WOGAN analyzers. We set $N = 50$ for all benchmarks except for $\BEAMNG_*$ for which we set $N = 20$ due to time constraints. The WOGAN algorithm used the benchmark input parameterization of \autoref{ssec:similarity_measures}, and it was set up as described in \autoref{sec:wogan_setup}.

For the WOGAN algorithm, we allocated an execution budget of $1500$ for the ARCH-COMP 2023 benchmarks and a budget of $300$ for the $\AMBIEGEN_*$ and $\BEAMNG_*$ benchmarks. For all benchmarks, we set the initial random budget $B_R$ to $75$ and the training delay $d_T$ to $3$. For the ARCH-COMP 2023 benchmarks, we select $\theta = 0.21$, so that with the execution budget of $1500$ on average $25 \%$ of the budget is used for uniform random sampling. The same is true for the other benchmarks as then $\theta = 0$ and $75 / 300 = 0.25$. The $\AMBIEGEN_*$ and the $\BEAMNG_*$ benchmarks get a smaller execution budget because the SUT
evaluation is slow in the $\BEAMNG_*$ benchmarks. We also wanted these benchmarks to have equal budgets as they
are similar to each other. In order to compute the metrics of \autoref{sec:evaluation_methodology}, we set the generator sample size $M$ also to $1500$ for the ARCH-COMP 2023 benchmarks and to $300$ for the remaining two. We remark that the budget $1500$ matches the SUT execution budget of
the ARCH-COMP 2023 competition.
We emphasize that each of the $N$ samples for WOGAN used a different WOGAN analyzer.

\begin{table}
\resizebox{\textwidth}{!}{
  \begin{tabular}{lrrrrrr}
  \toprule
       & \multicolumn{3}{c}{RANDOM}
       & \multicolumn{3}{c}{WOGAN} \\
  \midrule \midrule
  Req.
       & \multicolumn{1}{c}{$D$-score} & \multicolumn{2}{c}{$Q$-scores}
       & \multicolumn{1}{c}{$D$-score} & \multicolumn{2}{c}{$Q$-scores} \\
  \cmidrule(lr){1-1}
  \cmidrule(lr){2-2}
  \cmidrule(lr){3-4}
  \cmidrule(lr){5-5}
  \cmidrule(lr){6-7}
  $\textup{AFC27}$ &
  \worst 1.000 (0.00) &  
  \worst 0.097 (0.00) &  
  \worst 0.737 (0.01) &  
  \best 1.000 (0.00) &  
  \best 0.063 (0.01) &  
  \best 0.164 (0.03) \\ 
  \cmidrule(lr){1-1}
  \cmidrule(lr){2-2}
  \cmidrule(lr){3-4}
  \cmidrule(lr){5-5}
  \cmidrule(lr){6-7}
  $\textup{AT1}$ &
  \worst 1.000 (0.00) &  
  \worst 0.270 (0.00) &  
  \worst 0.445 (0.00) &  
  \best 0.999 (0.00) &  
  \best 0.099 (0.01) &  
  \best 0.130 (0.01) \\ 
  $\textup{AT6}_{4,35,3000}$ &
  \worst 1.000 (0.00) &  
  \worst 0.539 (0.01) &  
  \worst 0.892 (0.00) &  
  \best 1.000 (0.00) &  
  \best 0.002 (0.00) &  
  \best 0.086 (0.02) \\ 
  $\textup{AT6}_{8,50,3000}$ &
  \worst 1.000 (0.00) &  
  \worst 0.535 (0.01) &  
  \worst 0.892 (0.00) &  
  \best 1.000 (0.00) &  
  \best 0.031 (0.01) &  
  \best 0.121 (0.02) \\ 
  $\textup{AT6}_{20,65,3000}$ &
  \worst 1.000 (0.00) &  
  \worst 0.531 (0.01) &  
  \worst 0.892 (0.00) &  
  \best 1.000 (0.00) &  
  \best 0.032 (0.02) &  
  \best 0.127 (0.02) \\ 
  \cmidrule(lr){1-1}
  \cmidrule(lr){2-2}
  \cmidrule(lr){3-4}
  \cmidrule(lr){5-5}
  \cmidrule(lr){6-7}
  $\textup{CC3}$ &
  \worst 1.000 (0.00) &  
  \worst 0.012 (0.00) &  
  \worst 0.018 (0.00) &  
  \best 1.000 (0.00) &  
  \best 0.005 (0.00) &  
  \best 0.011 (0.00) \\ 
  $\textup{CC4}$ &
  \worst 1.000 (0.00) &  
  \worst 0.007 (0.00) &  
  \worst 0.007 (0.00) &  
  \best 1.000 (0.00) &  
  \best 0.006 (0.00) &  
  \best 0.007 (0.00) \\ 
  \cmidrule(lr){1-1}
  \cmidrule(lr){2-2}
  \cmidrule(lr){3-4}
  \cmidrule(lr){5-5}
  \cmidrule(lr){6-7}
  $\textup{F16}$ &
  \worst 0.922 (0.01) &  
  \worst 0.060 (0.00) &  
  \worst 0.113 (0.00) &  
  \best 0.520 (0.09) &  
  \best 0.046 (0.00) &  
  \best 0.066 (0.01) \\ 
  \midrule
  $\textup{NN}_{0.03}$ &
  \worst 0.996 (0.00) &  
  \worst 0.163 (0.00) &  
  \worst 0.461 (0.01) &  
  \best 0.663 (0.12) &  
  \best 0.084 (0.01) &  
  \best 0.161 (0.01) \\ 
  $\textup{NN}_{0.04}$ &
  \worst 0.996 (0.00) &  
  \worst 0.219 (0.00) &  
  \worst 0.488 (0.01) &  
  \best 0.576 (0.14) &  
  \best 0.147 (0.02) &  
  \best 0.204 (0.02) \\ 
  \cmidrule(lr){1-1}
  \cmidrule(lr){2-2}
  \cmidrule(lr){3-4}
  \cmidrule(lr){5-5}
  \cmidrule(lr){6-7}
  $\textup{AMBIEGEN}_5$ &
  \worst 0.165 (0.01) &  
  \worst 0.406 (0.01) &  
  \worst 0.551 (0.01) &  
  \best 0.111 (0.02) &  
  \best 0.244 (0.04) &  
  \best 0.418 (0.02) \\ 
  $\textup{AMBIEGEN}_7$ &
  \worst 0.300 (0.01) &  
  \worst 0.317 (0.01) &  
  \worst 0.493 (0.01) &  
  \best 0.247 (0.04) &  
  \best 0.184 (0.03) &  
  \best 0.356 (0.02) \\ 
  $\textup{AMBIEGEN}_9$ &
  \worst 0.407 (0.02) &  
  \worst 0.251 (0.01) &  
  \worst 0.430 (0.01) &  
  \best 0.362 (0.05) &  
  \best 0.159 (0.03) &  
  \best 0.339 (0.02) \\ 
  \cmidrule(lr){1-1}
  \cmidrule(lr){2-2}
  \cmidrule(lr){3-4}
  \cmidrule(lr){5-5}
  \cmidrule(lr){6-7}
  $\textup{BEAMNG}_5$ &
  \worst 0.166 (0.01) &  
  \worst 0.647 (0.02) &  
  \worst 0.905 (0.01) &  
  \best 0.131 (0.01) &  
  \best 0.295 (0.10) &  
  \best 0.681 (0.03) \\ 
  $\textup{BEAMNG}_7$ &
  \worst 0.294 (0.01) &  
  \worst 0.580 (0.01) &  
  \worst 0.812 (0.01) &  
  \best 0.220 (0.02) &  
  \best 0.334 (0.06) &  
  \best 0.627 (0.02) \\ 
  $\textup{BEAMNG}_9$ &
  \worst 0.366 (0.01) &  
  \worst 0.473 (0.02) &  
  \worst 0.739 (0.01) &  
  \best 0.307 (0.03) &  
  \best 0.267 (0.05) &  
  \best 0.590 (0.03) \\ 
\end{tabular}
}
\caption{Diversity scores ($D$-score) and quantile scores ($Q$-scores) for the generators of \autoref{ssec:rq1}. For each score, we report the mean and standard deviation in parenthesis. The statistics are computed over $50$ independent replicas except for the $\BEAMNG_*$ benchmark in which $20$ replicas are used. Test suite rank is indicated by color: \bestbox = best rank, \worstbox = worst rank.}\label{tbl:generator_quality}
\end{table}

Consider now the results of \autoref{tbl:generator_quality}. Evidently, WOGAN has better rank than RANDOM on all benchmarks, which means that WOGAN generators tend to have lower $Q$-scores, that is, they are capable of generating tests with lower robustness than uniform random sampling. On average WOGAN generators have small or negligible diversity loss on all benchmarks except $\FA$, $\NN_{0.03}$, $\NN_{0.04}$, $\AMBIEGEN_5$, $\BEAMNG_5$, and $\BEAMNG_7$.  First of all, it is expected that the set of low-robustness tests is small compared to the whole input space, so it is only natural that WOGAN suffers some diversity loss. Secondly, when the input space itself is small (in dimension), concentrating on a small set has even more dramatic effect to the diversity score. Indeed, when the benchmarks $\AMBIEGEN_*$ and $\BEAMNG_*$ are omitted, it can be computed that if $K$ mutually dissimilar tests have been sampled, the probability of uniformly randomly sampling another test that is dissimilar to all preceding tests is $1 - K \cdot 0.05^D$ where $D$ is the input space dimension. In the case of the benchmarks $\FA$, $\NN_{0.03}$, and $\NN_{0.04}$, we have $D = 3$, so the probability is only $0.8125$ when $K = 1500$. It is thus quite likely that even uniform random generators, which should have the greatest diversity score, generate similar tests. Judging from the low diversity scores of RANDOM for the $\AMBIEGEN_*$ and $\BEAMNG_*$ benchmarks, the same appears to be true for these benchmarks. We conclude that the WOGAN generators create diverse test suites: the tests are not collapsed to few nonvaried tests. This indicates that the features of the WOGAN algorithm that are designed to enable diversity do work as intended.

\subsubsection{Answer to RQ1}
WOGAN generators typically have minor diversity loss compared to uniform random generator, and they rank better. WOGAN generator robustness distributions are diverse distributions that concentrate closer to $0$ than the robustness distribution of a uniform random sampler.

\subsection{RQ2: Effects of WOGAN Features}\label{ssec:rq2}
RQ2 asks how the WGAN trained by the WOGAN algorithm performs when the proposed WGAN sampler and analyzer are altered and what is the effect of rejection sampling on WOGAN generator samples. For the latter question, we consider the following WOGAN generator variant.\\

\noindent
\textbf{WOGAN+NA} (WOGAN generator without a WOGAN analyzer). This generator simply samples a trained WOGAN generator and does not perform rejection sampling with the help of an analyzer. The comparison between WOGAN from \autoref{tbl:generator_quality} and WOGAN+NA illustrates how a WOGAN analyzer shapes a generator's distribution.\\

For the former question, we consider the following variants of the WOGAN algorithm.\\

\noindent
\textbf{WOGAN+RS}. This WOGAN variant replaces the WGAN training data sampler (line $4$ of \autoref{alg_sampler}) with a sampler that samples the training data uniformly randomly from the tests executed on the SUT. \\

\noindent
\textbf{WOGAN+RA}. This variant replaces the trained WOGAN analyzer with a random analyzer that estimates the robustness of a test by selecting a number in $[0,1]$ uniformly randomly. \\

\noindent
\textbf{WOGAN+PA}. Here the trained WOGAN analyzer is replaced by a perfect analyzer that
always estimates test robustness correctly. The perfect analyzer is implemented as a direct execution call to the SUT.
We do not count these additional executions against the execution budget.
\\

We follow the same setup as in \autoref{ssec:rq1} on the same benchmarks except we omitted the $\BEAMNG_*$ benchmarks due to time constraints. This yielded $50$ generators for each variant. The generators were evaluated as in \autoref{ssec:rq1}, and the results are presented in \autoref{tbl:features}. Notice that the generator evaluation is done on samples obtained without using the rejection sampling using an analyzer as described in \autoref{ssec:wgan_sampling}. This is because the research question concerns the effects of modification on the WGAN itself not on the rejection sampling method.

\subsubsection{Effect of the Rejection Sampling}\label{ssec:rq2_rejection_sampling}
\begin{figure*}
\centering
\includegraphics[trim=20 5 35 0,clip,width=0.41\textwidth]{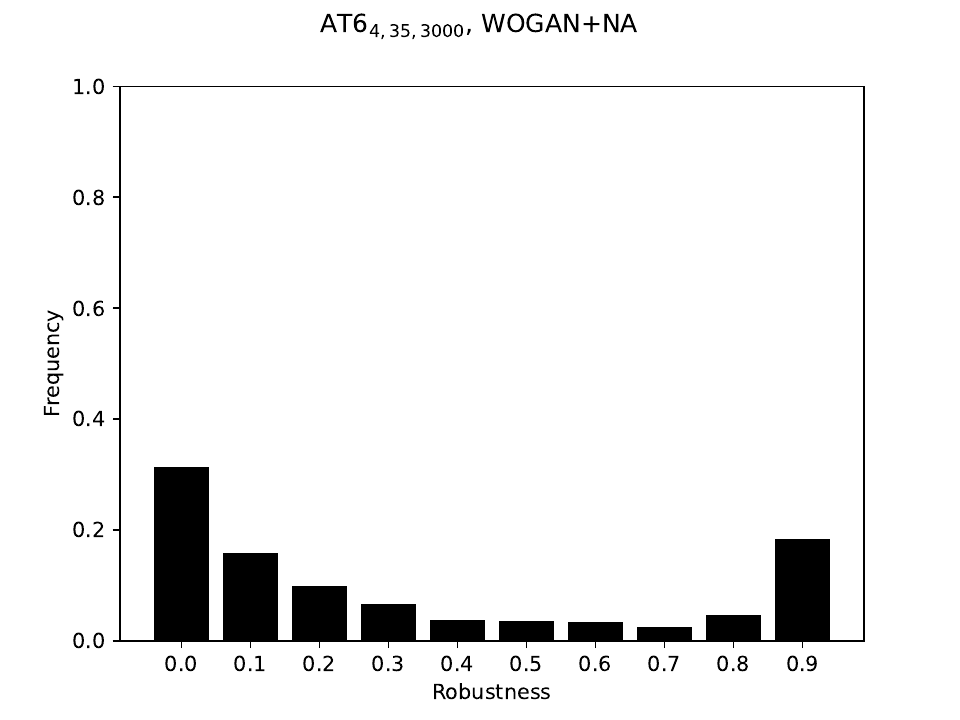}
\includegraphics[trim=20 5 35 0,clip,width=0.41\textwidth]{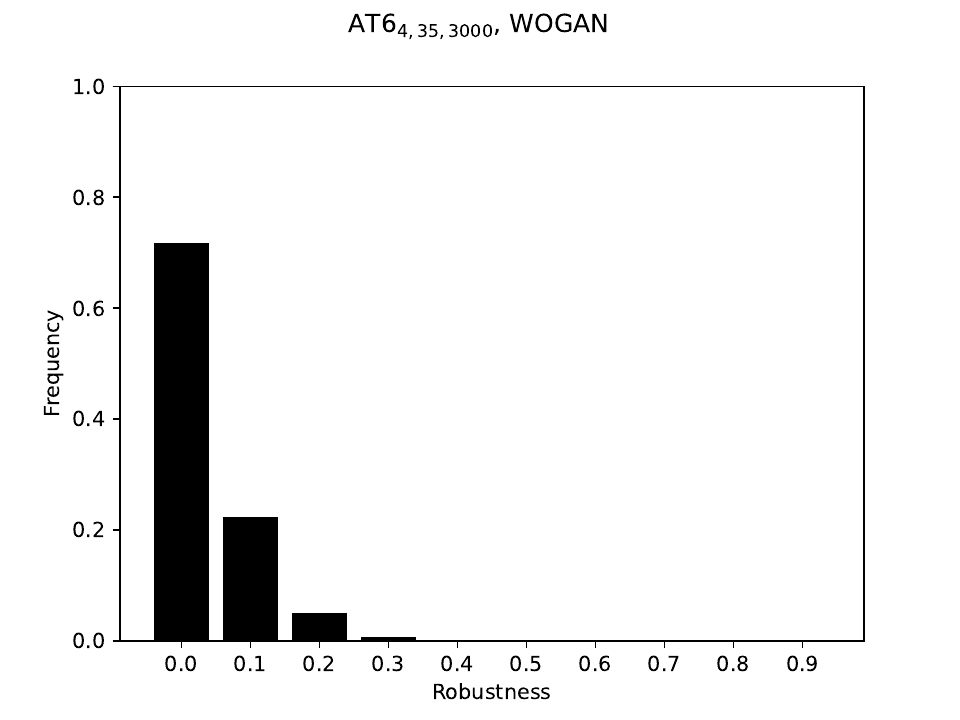}
\includegraphics[trim=20 5 35 0,clip,width=0.41\textwidth]{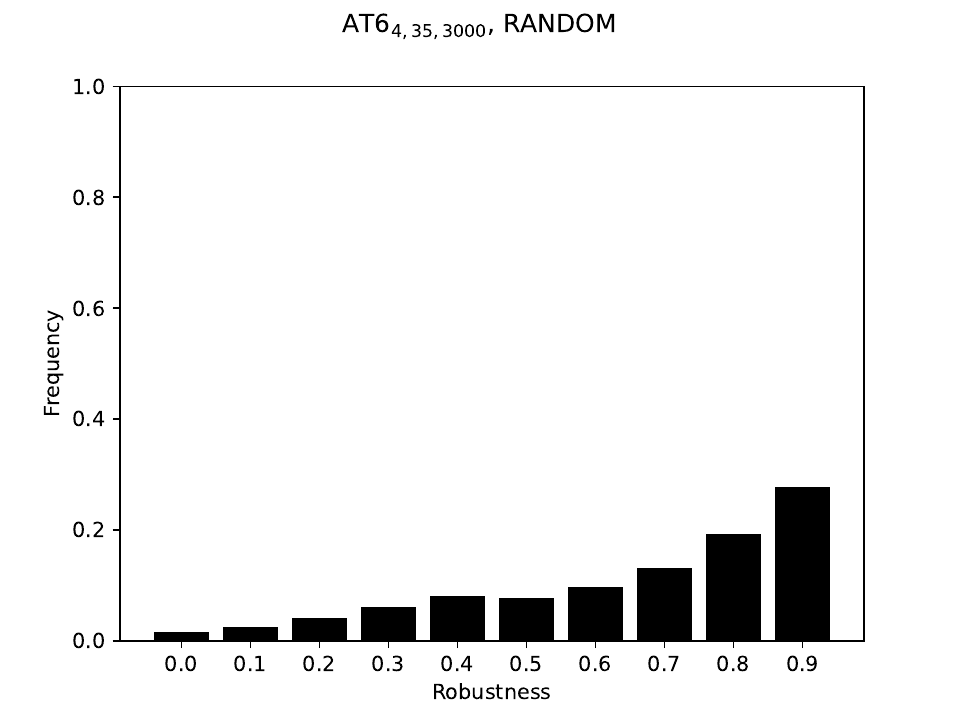}
\includegraphics[trim=20 5 35 0,clip,width=0.41\textwidth]{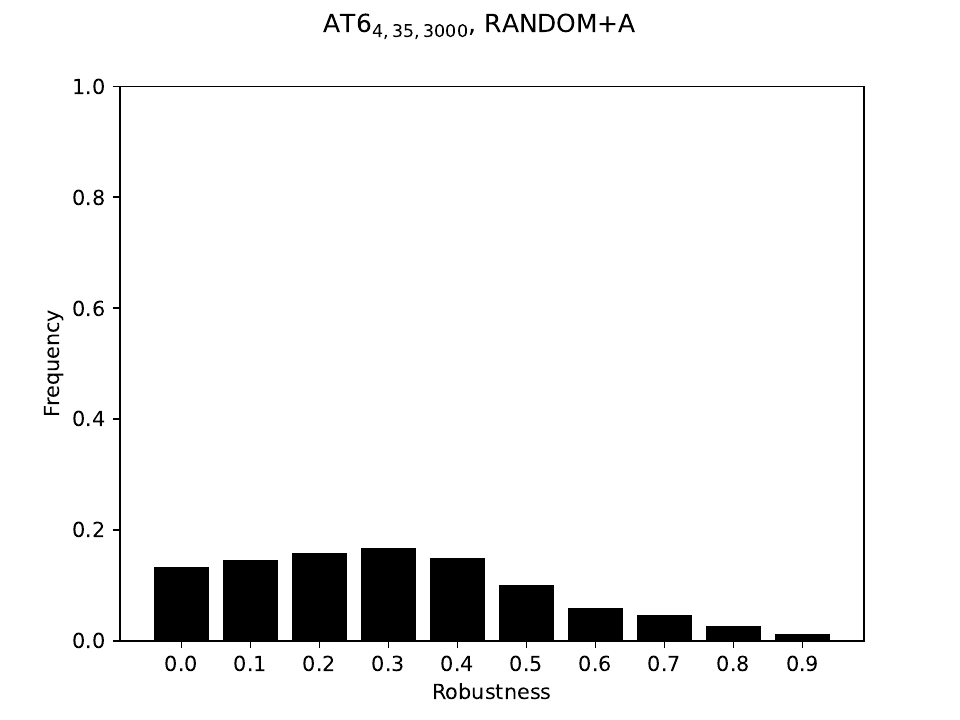}
\caption{Example robustness distributions in the $\ATVIA$ benchmark. RANDOM+A stands for a sample obtained by sampling the input space uniformly randomly and applying rejection sampling with a WOGAN analyzer.}\label{fig:distributions}
\end{figure*}

Before discussing the results of \autoref{tbl:features}, let us see how using a WOGAN analyzer in rejection sampling shapes a generator's distribution. \autoref{fig:distributions} shows that the robustness distribution of WOGAN+NA has a spike near $0.9$ and that the analyzer, as expected, has a tendency to thin the tail of a distribution. The same effect is present when shaping applying the rejection sampling to a uniform random generator. A further inspection of the distributions shows that this behavior is typical across all benchmarks. We do not know the reason for the spikes. We conjecture the cause to be an artifact of the way we train the WGAN. The training data always includes some high-robustness tests to maintain diversity, and this can be conflicting for the generator.

When comparing WOGAN+NA from \autoref{tbl:features} to WOGAN from \autoref{tbl:generator_quality}, we see that using the rejection sampling improves both $Q_U$-scores and $Q_L$-scores. The former are greatly improved, but the effect is not as dramatic for the latter. On the benchmarks $\CCIII$, $\CCIV$, and $\FA$, WOGAN and WOGAN+NA rank equally. We do not know the reason for this. The latter two benchmarks are hard to falsify for WOGAN, and this suggests that achieving low $Q$-scores is difficult. WOGAN+NA generators have small or negligible diversity loss on all benchmarks except $\FA$, $\NN_{0.03}$, $\NN_{0.04}$, and $\AMBIEGEN_5$. Thus they have less diversity loss than WOGAN generators. This is an expected result as the analyzer necessarily concentrates the sampling to a smaller subset of the input space.

The difference between WOGAN and WOGAN+NA shows that using an analyzer in the rejection sampling of the WGAN is crucial for best results: it moves the distribution closer to $0$ and removes unwanted distribution tails. The drawback is that using an analyzer incurs some diversity loss, but the effect is quite small.

\subsubsection{Effect of the WGAN Sampler}\label{ssec:rq2_sampler}
It is evident from \autoref{tbl:features} that WOGAN+RS generators lose in ranking to WOGAN+NA generators on all benchmarks except on $\FA$, $\NN_{0.03}$, $\NN_{0.04}$, and $\AMBIEGEN_*$ on which the rank is equal. We are not sure what causes this, but we believe it is related to the fact that the analyzer protects from degradation (see the next paragraph) and the fact that these benchmarks are difficult to learn a good distribution on. Indeed, at least on the benchmarks $\NN_{0.03}$, $\NN_{0.04}$, and $\AMBIEGEN_*$, WOGAN+NA achieves similar $Q_L$-scores as RANDOM (see \autoref{tbl:generator_quality}) indicating difficulty in learning how to generate low-robustness tests. Nevertheless we conclude that the proposed training data sampler is a good feature and generally improves the generators compared to a random sampler.

Interestingly WOGAN+RS generators beat RANDOM generators (see \autoref{tbl:generator_quality}) in rank except on the benchmarks $\CCIV$ and $\AMBIEGEN_*$ on which the rank is equal. This shows that the WGAN still learns a better distribution even when a subpar training data sampling is used. We think this is due to the analyzer: even if the training data is sampled randomly, the tests that are added to it are estimated to have low robustness by the analyzer. Thus, as long as the analyzer estimates better than a random guess, the training data will be biased towards low-robustness tests and fat distribution tails are removed. We take this as an indication that the analyzer protects against performance degradation when the sampler is not good. This protection is good as it means that the WOGAN algorithm is robust against minor sampler modifications. We conjecture that replacing the analyzer of RANDOM+RS with a random analyzer as in RANDOM+RA yields results similar to those of RANDOM from \autoref{tbl:generator_quality}, but we have not studied this.

Regarding diversity, we see that WOGAN+NA has lower $D$-score than WOGAN+RS on the benchmarks $\FA$, $\NN_{0.03}$, and $\NN_{0.04}$. We do not view the difference as large. Moreover, the discussion of \autoref{ssec:rq1} still applies: the input space is small in these benchmarks, so generators with even slightly lower $Q$-scores are likely to have a detectable change in the $D$-score. We conclude that altering the sampler does not lead to significant changes in the diversity.

\subsubsection{Effect of the Analyzer}
Let us then consider the effect of the analyzer. WOGAN+RA generators never rank better than WOGAN+NA generators, and they rank worse than WOGAN+NA on the benchmarks $\ATVIA$, $\ATVIB$, and $\ATVIC$. Surprisingly a perfect analyzer does not improve performance: WOGAN+NA and WOGAN+PA have equal rank. We conclude that the analyzer is a good feature: a reasonably well-tuned analyzer improves performance. The results also indicate that the analyzer training setup (see \autoref{ssec:analyzer_model}) is good: improving it does not seem to lead to significantly better performance.

In \autoref{ssec:rq2_sampler}, we argued that a well-tuned analyzer protects against performance degradation when the sampler is not good. Comparing the Q-scores of WOGAN+RA to RANDOM from \autoref{tbl:generator_quality} indicates that the converse is also true: the presence of a good sampler protects against performance degradation when the analyzer is of poor quality. Thus we conclude that having both a sampler and an analyzer is good as together they make the WOGAN algorithm robust against minor changes in these components' setup.

Regarding diversity there is not much to comment on. WOGAN+NA and WOGAN+PA have equal diversity scores for all practical purposes. There is a difference in $D$-score between WOGAN+NA and WOGAN+RA on the benchmarks $\FA$, $\NN_{0.03}$, and $\NN_{0.04}$, but the discussion of \autoref{ssec:rq2_sampler} still applies.

\begin{table}
\resizebox{\textwidth}{!}{
  \begin{tabular}{lrrrrrrrrrrrrrrr}
  \toprule
       & \multicolumn{3}{c}{WOGAN+NA}
       & \multicolumn{3}{c}{WOGAN+RS}
       & \multicolumn{3}{c}{WOGAN+RA}
       & \multicolumn{3}{c}{WOGAN+PA} \\
  \midrule \midrule
  Req.
       & \multicolumn{1}{c}{$D$-score} & \multicolumn{2}{c}{$Q$-scores}
       & \multicolumn{1}{c}{$D$-score} & \multicolumn{2}{c}{$Q$-scores}
       & \multicolumn{1}{c}{$D$-score} & \multicolumn{2}{c}{$Q$-scores}
       & \multicolumn{1}{c}{$D$-score} & \multicolumn{2}{c}{$Q$-scores} \\
  \cmidrule(lr){1-1}
  \cmidrule(lr){2-2}
  \cmidrule(lr){3-4}
  \cmidrule(lr){5-5}
  \cmidrule(lr){6-7}
  \cmidrule(lr){8-8}
  \cmidrule(lr){9-10}
  \cmidrule(lr){11-11}
  \cmidrule(lr){12-13}
  \cmidrule(lr){14-14}
  $\textup{AFC27}$ &
  \best 1.000 (0.00) &  
  \best 0.071 (0.01) &  
  \best 0.485 (0.13) &  
  \worst 1.000 (0.00) &  
  \worst 0.093 (0.02) &  
  \worst 0.531 (0.11) &  
  \best 1.000 (0.00) &  
  \best 0.075 (0.02) &  
  \best 0.591 (0.14) &  
  \best 1.000 (0.00) &  
  \best 0.062 (0.01) &  
  \best 0.674 (0.09) \\ 
  \cmidrule(lr){1-1}
  \cmidrule(lr){2-2}
  \cmidrule(lr){3-4}
  \cmidrule(lr){5-5}
  \cmidrule(lr){6-7}
  \cmidrule(lr){8-8}
  \cmidrule(lr){9-10}
  \cmidrule(lr){11-11}
  \cmidrule(lr){12-13}
  \cmidrule(lr){14-14}  
  $\textup{AT1}$ &
  \best 0.980 (0.01) &  
  \best 0.120 (0.02) &  
  \best 0.308 (0.04) &  
  \worst 1.000 (0.00) &  
  \worst 0.158 (0.02) &  
  \worst 0.345 (0.02) &  
  \best 0.996 (0.01) &  
  \best 0.122 (0.01) &  
  \best 0.328 (0.04) &  
  \best 0.980 (0.02) &  
  \best 0.117 (0.01) &  
  \best 0.307 (0.05) \\ 
  $\textup{AT6}_{4,35,3000}$ &
  \best 1.000 (0.01) &  
  \best 0.044 (0.02) &  
  \best 0.655 (0.08) &  
  \worst 1.000 (0.00) &  
  \worst 0.127 (0.05) &  
  \worst 0.804 (0.05) &  
  \worst 1.000 (0.00) &  
  \worst 0.158 (0.06) &  
  \worst 0.727 (0.06) &  
  \best 0.998 (0.00) &  
  \best 0.060 (0.02) &  
  \best 0.668 (0.06) \\ 
  $\textup{AT6}_{8,50,3000}$ &
  \best 0.999 (0.00) &  
  \best 0.090 (0.02) &  
  \best 0.696 (0.09) &  
  \worst 1.000 (0.00) &  
  \worst 0.189 (0.04) &  
  \worst 0.796 (0.06) &  
  \worst 1.000 (0.00) &  
  \worst 0.220 (0.06) &  
  \worst 0.715 (0.06) &  
  \best 0.999 (0.00) &  
  \best 0.120 (0.04) &  
  \best 0.694 (0.06) \\ 
  $\textup{AT6}_{20,65,3000}$ &
  \best 0.999 (0.00) &  
  \best 0.090 (0.02) &  
  \best 0.696 (0.10) &  
  \worst 1.00 (0.00) &  
  \worst 0.195 (0.04) &  
  \worst 0.789 (0.07) &  
  \worst 1.00 (0.00) &  
  \worst 0.20 (0.04) &  
  \worst 0.67 (0.07) &  
  \best 0.999 (0.00) &  
  \best 0.111 (0.03) &  
  \best 0.689 (0.09) \\ 
  \cmidrule(lr){1-1}
  \cmidrule(lr){2-2}
  \cmidrule(lr){3-4}
  \cmidrule(lr){5-5}
  \cmidrule(lr){6-7}
  \cmidrule(lr){8-8}
  \cmidrule(lr){9-10}
  \cmidrule(lr){11-11}
  \cmidrule(lr){12-13}
  \cmidrule(lr){14-14}  
  $\textup{CC3}$ &
  \best 1.000 (0.00) &  
  \best 0.005 (0.00) &  
  \best 0.011 (0.00) &  
  \worst 1.000 (0.00) &  
  \worst 0.005 (0.00) &  
  \worst 0.018 (0.00) &  
  \best 1.000 (0.00) &  
  \best 0.005 (0.00) &  
  \best 0.011 (0.00) &  
  \best 1.000 (0.00) &  
  \best 0.005 (0.00) &  
  \best 0.011 (0.00) \\ 
  $\textup{CC4}$ &
  \best 1.000 (0.00) &  
  \best 0.006 (0.00) &  
  \best 0.007 (0.00) &  
  \worst 1.000 (0.00) &  
  \worst 0.007 (0.00) &  
  \worst 0.007 (0.00) &  
  \best 1.000 (0.00) &  
  \best 0.007 (0.00) &  
  \best 0.007 (0.00) &  
  \best 1.000 (0.00) &  
  \best 0.007 (0.00) &  
  \best 0.007 (0.00) \\ 
  \cmidrule(lr){1-1}
  \cmidrule(lr){2-2}
  \cmidrule(lr){3-4}
  \cmidrule(lr){5-5}
  \cmidrule(lr){6-7}
  \cmidrule(lr){8-8}
  \cmidrule(lr){9-10}
  \cmidrule(lr){11-11}
  \cmidrule(lr){12-13}
  \cmidrule(lr){14-14}  
  $\textup{F16}$ &
  \best 0.567 (0.05) &  
  \best 0.048 (0.00) &  
  \best 0.076 (0.01) &  
  \best 0.734 (0.05) &  
  \best 0.052 (0.01) &  
  \best 0.083 (0.01) &  
  \best 0.599 (0.04) &  
  \best 0.048 (0.00) &  
  \best 0.078 (0.01) &  
  \best 0.559 (0.05) &  
  \best 0.048 (0.00) &  
  \best 0.076 (0.01) \\ 
  \cmidrule(lr){1-1}
  \cmidrule(lr){2-2}
  \cmidrule(lr){3-4}
  \cmidrule(lr){5-5}
  \cmidrule(lr){6-7}
  \cmidrule(lr){8-8}
  \cmidrule(lr){9-10}
  \cmidrule(lr){11-11}
  \cmidrule(lr){12-13}
  \cmidrule(lr){14-14}  
  $\textup{NN}_{0.03}$ &
  \best 0.841 (0.06) &  
  \best 0.119 (0.01) &  
  \best 0.231 (0.02) &  
  \best 0.936 (0.02) &  
  \best 0.147 (0.02) &  
  \best 0.247 (0.03) &  
  \best 0.927 (0.02) &  
  \best 0.138 (0.02) &  
  \best 0.227 (0.02) &  
  \best 0.868 (0.06) &  
  \best 0.120 (0.01) &  
  \best 0.228 (0.02) \\  
  $\textup{NN}_{0.04}$ &
  \best 0.820 (0.07) &  
  \best 0.183 (0.02) &  
  \best 0.298 (0.03) &  
  \best 0.936 (0.02) &  
  \best 0.210 (0.02) &  
  \best 0.298 (0.03) &  
  \best 0.914 (0.03) &  
  \best 0.199 (0.02) &  
  \best 0.280 (0.02) &  
  \best 0.826 (0.07) &  
  \best 0.185 (0.02) &  
  \best 0.300 (0.04) \\ 
  \cmidrule(lr){1-1}
  \cmidrule(lr){2-2}
  \cmidrule(lr){3-4}
  \cmidrule(lr){5-5}
  \cmidrule(lr){6-7}
  \cmidrule(lr){8-8}
  \cmidrule(lr){9-10}
  \cmidrule(lr){11-11}
  \cmidrule(lr){12-13}
  \cmidrule(lr){14-14}  
  $\textup{AMBIEGEN}_5$ &
  \best 0.146 (0.02) &  
  \best 0.244 (0.04) &  
  \best 0.418 (0.02) &  
  \best 0.154 (0.01) &  
  \best 0.378 (0.01) &  
  \best 0.536 (0.01) &  
  \best 0.149 (0.01) &  
  \best 0.386 (0.02) &  
  \best 0.545 (0.01) &  
  \best 0.144 (0.01) &  
  \best 0.379 (0.01) &  
  \best 0.541 (0.01) \\ 
  $\textup{AMBIEGEN}_7$ &
  \best 0.272 (0.03) &  
  \best 0.285 (0.02) &  
  \best 0.478 (0.02) &  
  \best 0.288 (0.03) &  
  \best 0.293 (0.01) &  
  \best 0.479 (0.01) &  
  \best 0.287 (0.03) &  
  \best 0.302 (0.02) &  
  \best 0.485 (0.02) &  
  \best 0.294 (0.03) &  
  \best 0.295 (0.01) &  
  \best 0.482 (0.01) \\ 
  $\textup{AMBIEGEN}_9$ &
  \best 0.377 (0.04) &  
  \best 0.234 (0.02) &  
  \best 0.415 (0.01) &  
  \best 0.387 (0.04) &  
  \best 0.239 (0.02) &  
  \best 0.417 (0.01) &  
  \best 0.381 (0.03) &  
  \best 0.243 (0.02) &  
  \best 0.426 (0.02) &  
  \best 0.383 (0.04) &  
  \best 0.237 (0.01) &  
  \best 0.419 (0.02)    
\end{tabular}
}
\caption{Diversity scores ($D$-score) and quantile scores ($Q$-scores) for the generators of \autoref{ssec:rq2}. For each score, we report the mean and standard deviation in parenthesis. The statistics are computed over $50$ independent replicas. Test suite rank is indicated by color: \bestbox = best rank, \secondbox = second rank, \thirdbox = third rank, \worstbox = worst rank.}\label{tbl:features}
\end{table}

\subsubsection{Answer to RQ2}
The analyzer used in the rejection sampling of WOGAN generators removes fat tails of the plain WGAN robustness distributions and improves the concentration towards $0$ with minor diversity loss. The proposed WGAN sampler and the analyzer have the intended effect of improving quantile scores without significant diversity loss when compared to generators trained on a random sampler and on a random analyzer. A well-tuned sampler protects against performance degradation in the presence of an analyzer of poor quality, and conversely a reasonably accurate analyzer protects against degradation when a subpar sampler is used. This shows that the performance of WOGAN generators is robust against minor variations in the sampler and analyzer.

\subsection{RQ3: Falsification}\label{ssec:rq3}
In this section, we study the falsification effectiveness and the capability of finding diverse falsifying tests of the RANDOM and WOGAN generators from \autoref{ssec:rq1}. We compare their performance to that of the state-of-the-art RF-algorithms from the ARCH-COMP 2023 competition \cite{ARCH23}. The evaluation methods are described in \autoref{ssec:falsification_capability}.

The evaluation results are displayed in \autoref{tbl:fr_generators}. This table displays a ranking of the generators that is based on the diversity score; best ranked generator $\generator$ has highest falsification diversity $D_F(\generator)$. The results of the ARCH-COMP 2023 competition are found in \autoref{tbl:arch23_results}.

\subsubsection{Falsification Effectiveness}
From the data of \autoref{tbl:fr_generators}, we see that the trend is that WOGAN generators achieve at least as good falsification rate as RANDOM generators. On the $\ATVIB$, $\ATVIC$, $\AMBIEGEN_5$, and $\AMBIEGEN_7$ benchmarks, WOGAN generators have significantly better falsification rate than RANDOM generators. On the $\FA$ benchmark, RANDOM generators achieve higher falsification rate, but we do not view this as a big difference as here RANDOM managed by chance to produce a single falsifying input in only $3$ out of $50$ replicas. The benchmark $\AFC$ is interesting as WOGAN generators lose significantly to RANDOM generators. The reason is that the analyzer is not accurate enough when the true robustness is close to $0$. Indeed, WOGAN+NA achieves a falsification rate of $0.96$ on this benchmark, so the bad performance is due to the generator sampling not due to poorly trained WGAN.

In summary, WOGAN generators have better falsification effectiveness than RANDOM generators provided that the analyzer predicts accurately for low-robustness tests.

\subsubsection{Falsification Diversity}
Consider next the benchmarks for which RANDOM generators achieve positive falsification diversity. With the exception of benchmarks $\AFC$ and $\FA$, WOGAN generators always achieve higher falsification diversity than RANDOM generators, and often with a large difference. For instance, on $\ATVIA$ and $\NN_{0.03}$, WOGAN generators achieve respectively $107.5$ and $2.3$ times higher score. On the benchmarks $\CCIII$, $\BEAMNG_5$, and $\BEAMNG_7$, the score is at most twice of that of RANDOM generators, and on other benchmarks it is higher. Again, the score is in favor of RANDOM on the $\AFC$ benchmark due to analyzer inaccuracy.

The conclusion is that typically WOGAN generators find a significantly higher number of faults than RANDOM generators. Recall that falsification diversity is not the mean number of found falsifying tests but the mean diversity of the falsifying tests as specified in \autoref{ssec:diversity}, so a sample of falsifying inputs from a WOGAN generator is more similar to the expected sample from a $D_0$-generator than a sample from a RANDOM generator.

\subsubsection{Comparison to RF-algorithms}
Let us finally compare WOGAN generators against the RF-algorithms from the ARCH-COMP 2023 competition. The algorithms are listed in \autoref{tbl:arch23_results}, and they are described in detail in the competition report \cite{ARCH23}. It should be noted that none of the algorithms aim to train a generator in the sense of \autoref{ssec:rfg_problem} or to find multiple and diverse falsifying inputs. We can still compare the falsification ratios (the execution budget of ARCH-COMP 2023 is $1500$ matching the WOGAN generator sample size). The comparison is not without issues. The results of \autoref{tbl:arch23_results} are based on only $10$ replicas while we have collected $50$ replicas. Moreover, no data is available for all RF-algorithms on all benchmarks.

Judging from the data, WOGAN generators have as good falsification rate as the RF-algorithms on the benchmarks $\ATVIA$, $\ATVIB$, $\ATVIC$, $\CCIII$, and $\NN_{0.03}$. On the difficult benchmark $\CCIV$, WOGAN generators are on par with the majority of the RF-algorithms. On $\FA$ and $\NN_{0.04}$, WOGAN generators have inferior falsification effectiveness, but the majority of the ARCH-COMP 2023 RF-algorithms do not report data for these two benchmarks. Again, WOGAN generators perform poorly on $\AFC$. Finally, on $\ATI$, WOGAN generators lose to most RF-algorithms.

Overall we find that WOGAN generators compare favorably to the RF-algorithms of the ARCH-COMP 2023 competition in terms of the falsification rate. They are not always able to falsify the requirement, but when they are, the falsification effectiveness is not far behind the ARCH COMP 2023 RF-algorithms.

\begin{table}
\centering
\begin{tabular}{lrrrrrr}
  \toprule
       & \multicolumn{2}{c}{RANDOM}
       & \multicolumn{2}{c}{WOGAN} \\
  \midrule \midrule
  Req.
       & \multicolumn{1}{c}{FR} & \multicolumn{1}{c}{$D_F$}
       & \multicolumn{1}{c}{FR} & \multicolumn{1}{c}{$D_F$} \\
  \cmidrule(r){1-1}
  \cmidrule(lr){2-3}
  \cmidrule(lr){4-5}
  $\textup{AFC27}$ &
  \best 0.90             &  
  \best 2.7 (1.5)        &  
  \worst 0.42            &  
  \worst 2.2 (4.8)       \\ 
  \cmidrule(r){1-1}
  \cmidrule(lr){2-3}
  \cmidrule(lr){4-5}
  $\textup{AT1}$ &
  \best 0.00         &  
  \best 0.0 (0.0)    &  
  \best 0.00         &  
  \best 0.0 (0.0)    \\ 
  $\textup{AT6}_{4,35,3000}$ &
  \worst 0.98            &  
  \worst 4.2 (1.9)       &  
  \best 1.00            &  
  \best 451.4 (118.3)   \\ 
  $\textup{AT6}_{8,50,3000}$ &
  \worst 0.74          &  
  \worst 1.2 (0.9)     &  
  \best 1.00            &  
  \best 135.1 (83.5)    \\ 
  $\textup{AT6}_{20,65,3000}$ &
  \worst 0.92            &  
  \worst 2.2 (1.3)       &  
  \best 1.00            &  
  \best 149.3 (86.4)    \\ 
  \cmidrule(r){1-1}
  \cmidrule(lr){2-3}
  \cmidrule(lr){4-5}
  $\textup{CC3}$ &
  \worst 1.00           &  
  \worst 57.6 (6.9)     &  
  \best 1.00           &  
  \best 100.7 (60.8)  \\ 
  $\textup{CC4}$ &
  \best 0.00         &  
  \best 0.0 (0.0)    &  
  \best 0.00         &  
  \best 0.0 (0.0)    \\ 
  \cmidrule(r){1-1}
  \cmidrule(lr){2-3}
  \cmidrule(lr){4-5}
   $\textup{F16}$ &
  \best 0.06             &  
  \best 0.1 (0.2)        &  
  \worst 0.00             &  
  \worst 0.0 (0.0)        \\ 
  \cmidrule(r){1-1}
  \cmidrule(lr){2-3}
  \cmidrule(lr){4-5}
  $\textup{NN}_{0.03}$ &
  \worst 1.00            &  
  \worst 11.0 (3.3)      &  
  \best 1.00            &  
  \best 25.8 (14.9)     \\ 
  $\textup{NN}_{0.04}$ &
  \best 0.00             &  
  \best 0.0 (0.0)        &  
  \best 0.02             &  
  \best 0.0 (0.1)        \\ 
  \cmidrule(r){1-1}
  \cmidrule(lr){2-3}
  \cmidrule(lr){4-5}
  $\textup{AMBIEGEN}_5$ &
  \worst 0.22            &  
  \worst 0.2 (0.4)       &  
  \best 0.88            &  
  \best 3.1 (1.8)       \\ 
  $\textup{AMBIEGEN}_7$ &
  \worst 0.64            &  
  \worst 1.1 (1.1)       &  
  \best 1.00            &  
  \best 7.0 (3.1)       \\ 
  $\textup{AMBIEGEN}_9$ &
  \worst 1.00          &  
  \worst 3.3 (1.5)     &  
  \best 1.00          &  
  \best 10.8 (4.4)    \\ 
  \cmidrule(r){1-1}
  \cmidrule(lr){2-3}
  \cmidrule(lr){4-5}
  $\textup{BEAMNG}^*_5$ &
  \worst 1.00          &  
  \worst 5.3 (1.9)     &  
  \best 1.00          &  
  \best 7.7 (1.4)     \\ 
  $\textup{BEAMNG}^*_7$ &
  \worst 1.00          &  
  \worst 3.8 (1.5)     &  
  \best 1.00          &  
  \best 5.8 (2.3)     \\ 
  $\textup{BEAMNG}^*_9$ &
  \worst 1.00          &  
  \worst 7.0 (1.9)     &  
  \best 1.00          &  
  \best 14.3 (3.1)       
\end{tabular}
\caption{Falsification results for WOGAN generators and uniform random generators. We report falsification rate (FR) and the falsifying diversity $D_F$ along with the corresponding standard deviation. The statistics are computed over $50$ independent replicas except for the $\BEAMNG_*$ benchmark in which $20$ replicas are used. The generators are ranked according to the numbers $D_F$: \bestbox = best rank, \worstbox = worst rank.}\label{tbl:fr_generators}
\end{table}

\begin{table}
\resizebox{\textwidth}{!}{
\begin{tabular}{lccccccc}
  \toprule
       & \multicolumn{1}{c}{ARIsTEO:}
       & \multicolumn{1}{c}{ATheNA:}
       & \multicolumn{1}{c}{FalCAuN:}
       & \multicolumn{1}{c}{FORESEE:}
       & \multicolumn{1}{c}{NNFal:}
       & \multicolumn{1}{c}{$\Psi$-TaLiRo:}
       & \multicolumn{1}{c}{STGEM:} \\
       & \multicolumn{1}{c}{ARX-2}
       & \multicolumn{1}{c}{}
       & \multicolumn{1}{c}{}
       & \multicolumn{1}{c}{}
       & \multicolumn{1}{c}{}
       & \multicolumn{1}{c}{ConBO}
       & \multicolumn{1}{c}{OGAN} \\
  \midrule \midrule
  Req.
       & \multicolumn{1}{c}{FR}
       & \multicolumn{1}{c}{FR}
       & \multicolumn{1}{c}{FR}
       & \multicolumn{1}{c}{FR}
       & \multicolumn{1}{c}{FR}
       & \multicolumn{1}{c}{FR}
       & \multicolumn{1}{c}{FR} \\
  \cmidrule(r){1-1}
  \cmidrule(lr){2-2}
  \cmidrule(lr){3-3}
  \cmidrule(lr){4-4}
  \cmidrule(lr){5-5}
  \cmidrule(lr){6-6}
  \cmidrule(lr){7-7}
  \cmidrule(lr){8-8}
  $\textup{AFC27}$ &
  $1.00$ &
  $0.80$ &
   &
  $1.00$ &
   &
  $1.00$ &
  $1.00$ \\
  \cmidrule(r){1-1}
  \cmidrule(lr){2-2}
  \cmidrule(lr){3-3}
  \cmidrule(lr){4-4}
  \cmidrule(lr){5-5}
  \cmidrule(lr){6-6}
  \cmidrule(lr){7-7}
  \cmidrule(lr){8-8}
  $\textup{AT1}_{20}$ &
  $0.00$ &
  $1.00$ &
  $1.00$ &
  $0.80$ &
  $0.20$ &
  $1.00$ &
  $1.00$ \\
  $\textup{AT6}_{4,35,3000}$ &
  $1.00$ &
  $1.00$ &
   &
  $1.00$ &
   &
  $0.90$ &
  $1.00$ \\
  $\textup{AT6}_{8,50,3000}$ &
  $1.00$ &
  $1.00$ &
   &
  $1.00$ &
   &
  $0.90$ &
  $1.00$ \\
  $\textup{AT6}_{20,65,3000}$ &
  $1.00$ &
  $1.00$ &
  $1.00$ &
  $1.00$ &
  $0.10$ &
  $0.70$ &
  $1.00$ \\
  \cmidrule(r){1-1}
  \cmidrule(lr){2-2}
  \cmidrule(lr){3-3}
  \cmidrule(lr){4-4}
  \cmidrule(lr){5-5}
  \cmidrule(lr){6-6}
  \cmidrule(lr){7-7}
  \cmidrule(lr){8-8}
  $\textup{CC3}$ &
  $1.00$ &
  $1.00$ &
   &
  $1.00$ &
   &
  $1.00$ &
  $1.00$ \\
  $\textup{CC4}$ &
  $0.00$ &
  $0.10$ &
   &
  $0.90$ &
   &
  $0.10$ &
  $0.00$ \\
  \cmidrule(r){1-1}
  \cmidrule(lr){2-2}
  \cmidrule(lr){3-3}
  \cmidrule(lr){4-4}
  \cmidrule(lr){5-5}
  \cmidrule(lr){6-6}
  \cmidrule(lr){7-7}
  \cmidrule(lr){8-8}
  $\textup{F16}$ &
   &
  $1.00$ &
   &
   &
   &
   &
  $0.70$ \\
  \cmidrule(r){1-1}
  \cmidrule(lr){2-2}
  \cmidrule(lr){3-3}
  \cmidrule(lr){4-4}
  \cmidrule(lr){5-5}
  \cmidrule(lr){6-6}
  \cmidrule(lr){7-7}
  \cmidrule(lr){8-8}
  $\textup{NN}_{0.03}$ &
  $1.00$ &
  $1.00$ &
   &
  $1.00$ &
   &
  $1.00$ &
  $1.00$ \\
  $\textup{NN}_{0.04}$ &
  $0.50$ &
  $0.10$ &
  &
  &
  &
  &
  $0.70$ \\
  \cmidrule(r){1-1}
  \cmidrule(lr){2-2}
  \cmidrule(lr){3-3}
  \cmidrule(lr){4-4}
  \cmidrule(lr){5-5}
  \cmidrule(lr){6-6}
  \cmidrule(lr){7-7}
  \cmidrule(lr){8-8}
\end{tabular}
}
\caption{Falsification rates over $10$ independent replicas for ARCH-COMP 2023 tools.}\label{tbl:arch23_results}
\end{table}

\subsubsection{Answer to RQ3}
Compared to uniform random generators, WOGAN generators are as effective, and typically more effective, for falsification. Care needs to be taken to ensure that the analyzer is accurate on low-robustness tests as inaccuracy can degrade the effectiveness. WOGAN generators find significantly more diverse falsifying tests than uniform random search: their samples of falsifying inputs are more similar to the expected samples from a $D_0$-generator than samples from uniform random sampling. WOGAN generators compare favorably to state-of-the-art RF-algorithms in terms of falsification effectiveness.

\subsection{RQ4: Computational Overhead}\label{ssec:rq5}
In order to see what is the computational cost of the WOGAN algorithm and test generation using a WOGAN generator, we have recorded the following statistics: the time $t_G$ it takes to generate a single test, the time $t_T$ it takes to train the models (the WGAN and the analyzer) during a single execution of the main loop of the WOGAN algorithm, the time $t_E$ it takes to execute a test on the SUT, the total $T$ to complete a replica, and the ratio $R$ of total time used for test execution over the total time. All times are measured in wall time and in seconds.

We have collected the means $\overline{t}_G$, $\overline{t}_T$, $\overline{t}_E$, and $\overline{T}$ and the ratios $R$ in Tables~\ref{tbl:time_algorithms},\ref{tbl:time_generators} for the computations performed for the results of \autoref{ssec:rq1}. This includes the time used to run the WOGAN algorithm to obtain the WOGAN generators (\autoref{tbl:time_algorithms}) and to sample the RANDOM and WOGAN generators (\autoref{tbl:time_generators}). We have observed random and unexplained slowdowns with the BeamNG.tech simulator, and this explains the variation in the column $\overline{T}$ of \autoref{tbl:time_algorithms} for the $\BEAMNG_*$ benchmarks.

\begin{table}
\centering
\small
\begin{tabular}{lrrrrr}
  \toprule
       & \multicolumn{5}{c}{WOGAN ALGORITHM} \\
  \midrule \midrule
  Req.
       & \multicolumn{1}{c}{$\overline{t}_G$} & \multicolumn{1}{c}{$\overline{t}_T$} & \multicolumn{1}{c}{$\overline{t}_E$} & \multicolumn{1}{c}{$\overline{T}$} & \multicolumn{1}{c}{$R$} \\
  \cmidrule(r){1-1}
  \cmidrule(lr){2-6}
  $\textup{AFC27}$ &
  $0.00299$ &
  $0.03254$ &
  $1.8$ &
  $2689.1$ &
  $0.98$ \\
  \cmidrule(r){1-1}
  \cmidrule(lr){2-6}
  $\textup{AT1}$ &
  $0.00378$ &
  $0.03902$ &
  $0.3$ &
  $497.6$ &
  $0.88$ \\
  $\textup{AT6}_{4,35,3000}$ &
  $0.00419$ &
  $0.04555$ &
  $0.3$ &
  $493.0$ &
  $0.86$ \\
  $\textup{AT6}_{8,50,3000}$ &
  $0.00430$ &
  $0.03855$ &
  $0.3$ &
  $511.4$ &
  $0.88$ \\
  $\textup{AT6}_{20,65,3000}$ &
  $0.00428$ &
  $0.03831$ &
  $0.3$ &
  $504.3$ &
  $0.88$ \\
  \cmidrule(r){1-1}
  \cmidrule(lr){2-6}
  $\textup{CC3}$ &
  $0.00184$ &
  $0.04724$ &
  $1.0$ &
  $1451.6$ &
  $0.95$ \\
  $\textup{CC4}$ &
  $0.00201$ &
  $0.05451$ &
  $1.0$ &
  $1510.9$ &
  $0.95$ \\
  \cmidrule(r){1-1}
  \cmidrule(lr){2-6}
  $\textup{F16}$ &
  $0.00115$ &
  $0.02601$ &
  $0.1$ &
  $137.9$ &
  $0.70$ \\
  \cmidrule(r){1-1}
  \cmidrule(lr){2-6}
  $\textup{NN}_{0.03}$ &
  $0.00265$ &
  $0.02985$ &
  $1.0$ &
  $1527.5$ &
  $0.97$ \\
  $\textup{NN}_{0.04}$ &
  $0.00329$ &
  $0.03035$ &
  $1.0$ &
  $1525.4$ &
  $0.97$ \\
  \cmidrule(r){1-1}
  \cmidrule(lr){2-6}
  $\textup{AMBIEGEN}_5$ &
  $0.01192$ &
  $0.07204$ &
  $0.0$ &
  $25.5$ &
  $0.04$ \\
  $\textup{AMBIEGEN}_7$ &
  $0.01239$ &
  $0.07208$ &
  $0.0$ &
  $28.7$ &
  $0.04$ \\
  $\textup{AMBIEGEN}_9$ &
  $0.01349$ &
  $0.06432$ &
  $0.0$ &
  $22.9$ &
  $0.06$ \\
  \cmidrule(r){1-1}
  \cmidrule(lr){2-6}
  $\textup{BEAMNG}_5$ &
  $2.66567$ &
  $0.01666$ &
  $28.1$ &
  $14132.6$ &
  $0.91$ \\
  $\textup{BEAMNG}_7$ &
  $5.94203$ &
  $0.01661$ &
  $29.4$ &
  $10714.1$ &
  $0.83$ \\
  $\textup{BEAMNG}_9$ &
  $8.95063$ &
  $0.01665$ &
  $31.6$ &
  $11937.7$ &
  $0.78$
\end{tabular}
\caption{Time measurements over $50$ independent replicas ($20$ for the $\BEAMNG_*$ benchmark) for executing the WOGAN algorithm. We report the means for the test generation time $t_G$, model training time $t_T$, test execution time $t_E$, and the total time $T$ used per replica. Additionally, we report the ratio $R$ of total execution time to the sum of total times. All times are measured in wall time and in seconds.}\label{tbl:time_algorithms}
\end{table}

\begin{table}
\centering
\small
\begin{tabular}{lrrrrrr}
  \toprule
       & \multicolumn{3}{c}{RANDOM}
       & \multicolumn{3}{c}{WOGAN} \\
  \midrule \midrule
  Req.
       & \multicolumn{1}{c}{$\overline{t}_G$} & \multicolumn{1}{c}{$\overline{T}$} & \multicolumn{1}{c}{$R$}
       & \multicolumn{1}{c}{$\overline{t}_G$} & \multicolumn{1}{c}{$\overline{T}$} & \multicolumn{1}{c}{$R$} \\
  \cmidrule(r){1-1}
  \cmidrule(lr){2-4}
  \cmidrule(lr){5-7}
  $\textup{AFC27}$ &
  $0.00003$ &
  $2618.8$ &
  $1.00$ &
  $0.00270$ &
  $2688.6$ &
  $1.00$ \\
  \cmidrule(r){1-1}
  \cmidrule(lr){2-4}
  \cmidrule(lr){5-7}
  $\textup{AT1}$ &
  $0.00003$ &
  $448.9$ &
  $1.00$ &
  $0.00319$ &
  $495.8$ &
  $0.99$ \\
  $\textup{AT6}_{4,35,3000}$ &
  $0.00003$ &
  $453.9$ &
  $1.00$ &
  $0.00263$ &
  $467.1$ &
  $0.99$ \\
  $\textup{AT6}_{8,50,3000}$ &
  $0.00003$ &
  $443.4$ &
  $1.00$ &
  $0.00283$ &
  $473.5$ &
  $0.99$ \\
  $\textup{AT6}_{20,65,3000}$ &
  $0.00003$ &
  $459.5$ &
  $1.00$ &
  $0.00281$ &
  $467.1$ &
  $0.99$ \\
  \cmidrule(r){1-1}
  \cmidrule(lr){2-4}
  \cmidrule(lr){5-7}
  $\textup{CC3}$ &
  $0.00003$ &
  $1434.1$ &
  $1.00$ &
  $0.00260$ &
  $1437.9$ &
  $1.00$ \\
  $\textup{CC4}$ &
  $0.00003$ &
  $1477.0$ &
  $1.00$ &
  $0.00263$ &
  $1507.7$ &
  $1.00$ \\
  \cmidrule(r){1-1}
  \cmidrule(lr){2-4}
  \cmidrule(lr){5-7}
  $\textup{F16}$ &
  $0.00002$ &
  $93.5$ &
  $1.00$ &
  $0.00109$ &
  $74.0$ &
  $0.98$ \\
  \cmidrule(r){1-1}
  \cmidrule(lr){2-4}
  \cmidrule(lr){5-7}
  $\textup{NN}_{0.03}$ &
  $0.00003$ &
  $1485.8$ &
  $1.00$ &
  $0.00263$ &
  $1507.4$ &
  $1.00$ \\
  $\textup{NN}_{0.04}$ &
  $0.00003$ &
  $1449.5$ &
  $1.00$ &
  $0.00345$ &
  $1535.2$ &
  $1.00$ \\
  \cmidrule(r){1-1}
  \cmidrule(lr){2-4}
  \cmidrule(lr){5-7}
  $\textup{AMBIEGEN}_5$ &
  $0.00087$ &
  $0.9$ &
  $0.74$ &
  $0.01081$ &
  $3.5$ &
  $0.16$ \\
  $\textup{AMBIEGEN}_7$ &
  $0.00127$ &
  $1.3$ &
  $0.77$ &
  $0.01076$ &
  $3.5$ &
  $0.19$ \\
  $\textup{AMBIEGEN}_9$ &
  $0.00178$ &
  $1.6$ &
  $0.74$ &
  $0.01184$ &
  $4.1$ &
  $0.21$ \\
  \cmidrule(r){1-1}
  \cmidrule(lr){2-4}
  \cmidrule(lr){5-7}
  $\textup{BEAMNG}_5$ &
  $0.24822$ &
  $8290.3$ &
  $0.99$ &
  $2.99526$ &
  $8004.9$ &
  $0.89$ \\
  $\textup{BEAMNG}_7$ &
  $0.96928$ &
  $11059.8$ &
  $0.97$ &
  $7.01121$ &
  $10307.1$ &
  $0.79$ \\
  $\textup{BEAMNG}_9$ &
  $1.03529$ &
  $9785.7$ &
  $0.97$ &
  $10.79897$ &
  $11717.8$ &
  $0.73$
\end{tabular}
\caption{Time measurements over $50$ independent replicas ($20$ for the $\BEAMNG_*$ benchmark) for RANDOM and WOGAN generators. We report the mean test generation time $\overline{t}_G$, the mean total time $\overline{T}$ used per replica, and the ratio $R$ of total execution time to the sum of total times. All times are measured in wall time and in seconds.}\label{tbl:time_generators}
\end{table}

\subsubsection{WOGAN Algorithm Computational Overhead}
It is immediate from \autoref{tbl:time_algorithms} that the mean training time $\overline{t}_T$ is one order of magnitude larger than the mean test generation time $\overline{t}_G$, but both are rather small as always $\overline{t}_T < 0.08$. With the exception of the $\AMBIEGEN_*$ benchmarks, the mean test execution time $\overline{t}_E$ is at least one order of magnitude larger than $\overline{t}_T$, so the WOGAN algorithm spends most time in test execution as indicated by the ratios $R$. Thus when $\overline{t}_E$ is at least $0.1$, we can estimate $\overline{T}$ by the number $B \cdot \overline{t}_E$, where $B$ is the test execution budget. In other words, the total time used per replica is proportional to the test execution budget $B$. When test execution is fast, as in the $\AMBIEGEN_*$ benchmarks, the used time resources are not high: on average $\AMBIEGEN_*$ replicas finish in less than half minute. We remark that the higher $\overline{t}_G$ for the $\AMBIEGEN_*$ and $\BEAMNG_*$ benchmarks is explained by the presence of invalid tests: sampling needs to continue until a valid test is found. The high numbers $\overline{t}_G$, especially in the case of $\BEAMNG_9$ benchmark, indicate that WOGAN has some difficulties in generating valid tests. We leave improving this aspect to future research.

We remark that the model training and WGAN sampling also depend on the size of the machine learning models. In this paper, the model sizes do not vary between benchmarks, and the models are overall rather small. Larger models might require significantly more computational resources, but we have not attempted to study this as the models in this paper already gave good results.

\subsubsection{Generator Sampling Computational Overhead}
We see from \autoref{tbl:time_generators} that, similar to above, the creation of a sample of size $B$ is largely determined by the number $B \cdot \overline{t}_E$ (the mean execution times do not meaningfully vary between the two tables). As expected, the generation time for RANDOM generators is negligible. The generation time is higher for WOGAN generators as the WGAN needs to be sampled and the analyzer used for robustness estimation. This requires two magnitudes more time, but this hardly changes the ratio except for the $\AMBIEGEN_*$ benchmark in which the extra requirement of sampling valid tests increases the time requirement even more. As remarked above, WOGAN generators have difficulties in generating valid tests on the $\BEAMNG_*$ benchmarks, but they always succeed eventually. Even RANDOM generators have this difficulty although it is not as marked.

\subsubsection{Answer to RQ4}
Most of the time used by the WOGAN algorithm is taken by the test execution when the execution of a test takes at least a tenth of a second. The time total time used is almost directly proportional to the mean test execution time and the test execution budget. The same is true when tests are generated using a WOGAN generator: the generation time is largely negligible. The use of larger models increases the needed time resources, but determining the effect requires a further study.

\section{Conclusions}\label{sec:conclusions}
We have described in detail the WOGAN algorithm that trains a generator that is an approximate solution to the RFG-problem of training a generator that can sample from the uniform distribution supported on the set of falsifying tests of a system under test (SUT). The WOGAN algorithm can be used for any deterministic black-box SUT that has outputs modeled as signals and requirements given in signal temporal logic (STL). The training is performed tabula rasa without any prior model or dataset on the SUT. To our knowledge, the RFG-problem has not been identified before in the context of validating cyber-physical systems, and WOGAN is the first algorithm to approximate a solution to this problem.

The generator trained by the WOGAN algorithm is modeled as a Wasserstein generative adversarial network (WGAN). At each iteration of the algorithm, a test is sampled from the partially trained WGAN with a rejection sampling method using a concurrently trained analyzer model, which estimates the STL robustness of system inputs. The sampled test is executed on the actual SUT, and the resulting system trace along with its robustness is added to the training data of the WGAN for further training. The WGAN training data shifts over time towards low-robustness tests according to the proposed quantile sampler. This shift achieves the online training: the initial uniformly randomly sampled training data is shifted towards low-robustness tests over time with the help of the partially trained WGAN.

In addition to presenting the WOGAN algorithm, we have formally defined the RFG-problem and proposed evaluation principles according to which it is possible to evaluate how well a given generator approximates a solution to the RFG-problem. The evaluation principles focus on two key aspects: i) the input diversity of the tests sampled from the generator and ii) the shape of the robustness distribution of the generator. In order to compute the input diversity, we have proposed a test clustering method that clusters tests according to a given test similarity metric; more clusters of different tests implies higher input diversity. We have proposed to capture the shape of the robustness distribution into two distribution quantiles that are estimated from a generator sample. The closer the quantiles are to zero, the better the generator is in generating low-robustness tests.

We have evaluated the WOGAN generators trained by the WOGAN algorithm according to the proposed evaluation criteria. The results indicate that the WOGAN generators approximate better a solution to the RFG-problem than uniform random generators. The WOGAN generators' robustness distributions contain tests of much lower robustness than those of uniform random generators. The input diversity score for WOGAN generators is typically close to that of uniform random generators indicating that the generated samples are input-diverse.

In addition to comparing WOGAN generators to uniform random generators, we have compared them to generators obtained from variants of the WOGAN algorithm itself. The results show that the WOGAN algorithm is robust against minor variations in its design: minor variations in analyzer accuracy or in the design of the training data sampler do not affect the generators' performance significantly. We consider the design of WOGAN sound: the designs yield the intended effects.

We evaluated WOGAN generator samples also in terms of falsifying tests they contain. We found that they contain significantly more input-diverse falsifying tests that uniform random samples. In terms of falsification effectiveness, we found that WOGAN generators achieve falsification rates that are on par with the state-of-the-art requirement falsification algorithms from the ARCH-COMP competition.

\section{Acknowledgments}
This research has received funding from the ECSEL Joint Undertaking (JU) under grant agreement No 101007350. The JU receives support from the European Union’s Horizon 2021 research and innovation program and Sweden, Austria, Czech Republic, Finland, France, Italy, Spain.

\bibliography{wogan}

\begin{thebibliography}{10}

\bibitem{2017:introduction_to_software_testing}
P.~Amman and J.~Offutt.
\newblock {\em Introduction to Software Testing}.
\newblock Cambridge University Press, 2017.
\newblock 2nd edition.

\bibitem{2011:staliro_a_tool_for_temporal_logic_falsification}
Y.~Annapureddy, C.~Liu, G.~Fainekos, and S.~Sankaranarayanan.
\newblock S-{T}a{L}i{R}o: {A} tool for temporal logic falsification for hybrid
  systems.
\newblock In P.~A. Abdulla and K.~R.~M. Leino, editors, {\em International
  Conference on Tools and Algorithms for the Construction and Analysis of
  Systems. {TACAS} 2011}, pages 254--257. Springer, 2011.

\bibitem{2019:multi_objective_search_for_effective_testing}
H.~Araujo, G.~Carvalho, M.~R. Mousavi, and A.~Sampaio.
\newblock Multi-objective search for effective testing of cyber-physical
  systems.
\newblock In P.~C. {\"O}lveczky and G.~Sala{\"u}n, editors, {\em Software
  Engineering and Formal Methods}, pages 183--202. Springer, 2019.
\newblock \href {https://doi.org/10.1007/978-3-030-30446-1_10}
  {\path{doi:10.1007/978-3-030-30446-1_10}}.

\bibitem{wgan}
M.~Arjovsky, S.~Chintala, and L.~Bottou.
\newblock {W}asserstein generative adversarial networks.
\newblock In {\em Proceedings of the 34th International Conference on Machine
  Learning}, volume~70 of {\em Proceedings of Machine Learning Research}, pages
  214--223. PMLR, 2017.
\newblock URL: \url{https://proceedings.mlr.press/v70/arjovsky17a.html}.

\bibitem{2024:falsification_using_reachability_of_surrogate_koopman}
S.~Bak, S.~Bogomolov, A.~Hekal, N.~Kochdumper, E.~Lew, A.~Mata, and A.~Rahmati.
\newblock Falsification using reachability of surrogate {K}oopman models.
\newblock In {\em Proceedings of the 27th ACM International Conference on
  Hybrid Systems: Computation and Control}. ACM, 2024.
\newblock \href {https://doi.org/10.1145/3641513.3650141}
  {\path{doi:10.1145/3641513.3650141}}.

\bibitem{SBFT-toolcomp24}
M.~Biagiola and S.~Klikovits.
\newblock {SBFT} tool competition 2024 - {C}yber-physical systems track.
\newblock In {\em 17th {ACM/IEEE} International Workshop on Search-Based and
  Fuzz Testing. {SBFT} 2024}, page 33–36. ACM, 2024.
\newblock \href {https://doi.org/10.1145/3643659.3643932}
  {\path{doi:10.1145/3643659.3643932}}.

\bibitem{SBFT-toolcomp23}
M.~Biagiola, S.~Klikovits, J.~Peltom\"{a}ki, and V.~Riccio.
\newblock {SBFT} tool competition 2023 - {C}yber-physical systems track.
\newblock In {\em 16th {IEEE/ACM} International Workshop on Search-Based and
  Fuzz Testing. {SBFT} 2023}, pages 45--48. {IEEE}, 2023.
\newblock \href {https://doi.org/10.1109/SBFT59156.2023.00010}
  {\path{doi:10.1109/SBFT59156.2023.00010}}.

\bibitem{2024:diversity_guided_search_exploration_for_self}
T.~Blattner, C.~Birchler, T.~Kehrer, and S.~Panichella.
\newblock Diversity-guided search exploration for self-driving cars test
  generation through {F}renet space encoding.
\newblock In {\em 17th {ACM/IEEE} International Workshop on Search-Based and
  Fuzz Testing. {SBFT} 2024}, pages 9--12. {ACM}, 2024.
\newblock \href {https://doi.org/10.1145/3643659.3643926}
  {\path{doi:10.1145/3643659.3643926}}.

\bibitem{frenetic}
E.~Castellano et~al.
\newblock Frenetic at the {SBST} 2021 tool competition.
\newblock In {\em 14th {IEEE/ACM} International Workshop on Search-Based
  Software Testing. {SBST} 2021}, pages 36--37, 2021.
\newblock \href {https://doi.org/10.1109/SBST52555.2021.00016}
  {\path{doi:10.1109/SBST52555.2021.00016}}.

\bibitem{2004:adaptive_random_testing}
T.~Y. Chen, H.~Leung, and I.~K. Mak.
\newblock Adaptive random testing.
\newblock In M.~J. Maher, editor, {\em Advances in Computer Science - {ASIAN}
  2004. Higher-Level Decision Making}, pages 320--329. Springer, 2004.
\newblock \href {https://doi.org/10.1007/978-3-540-30502-6_23}
  {\path{doi:10.1007/978-3-540-30502-6_23}}.

\bibitem{DBLP:journals/jair/CorsoMKLK21}
A.~Corso, R.~J. Moss, M.~Koren, R.~Lee, and M.~J. Kochenderfer.
\newblock A survey of algorithms for black-box safety validation of
  cyber-physical systems.
\newblock {\em J. Artif. Intell. Res.}, 72:377--428, 2021.
\newblock \href {https://doi.org/10.1613/jair.1.12716}
  {\path{doi:10.1613/jair.1.12716}}.

\bibitem{donze2010robust}
A.~Donz{\'e} and O.~Maler.
\newblock Robust satisfaction of temporal logic over real-valued signals.
\newblock In K.~Chatterjee and T.~A. Henzinger, editors, {\em Formal Modeling
  and Analysis of Timed Systems. FORMATS 2010}, volume 6246 of {\em Lecture
  Notes in Computer Science}, pages 92--106. Springer, 2010.
\newblock \href {https://doi.org/10.1007/978-3-642-15297-9_9}
  {\path{doi:10.1007/978-3-642-15297-9_9}}.

\bibitem{2023:a_survey_of_the_metrics_uses_and_subjects_of_diversity}
I.~T. Elgendy, R.~M. Hierons, and P.~McMinn.
\newblock A survey of the metrics, uses, and subjects of diversity-based
  techniques in software testing.
\newblock 2023.
\newblock Preprint.
\newblock \href {https://arxiv.org/abs/2311.09714} {\path{arXiv:2311.09714}}.

\bibitem{ARCH21}
G.~Ernst et~al.
\newblock {ARCH-COMP} 2021 category report: {F}alsification with validation of
  results.
\newblock In G.~Frehse and M.~Althoff, editors, {\em 8th International Workshop
  on Applied Verification of Continuous and Hybrid Systems. {ARCH21}},
  volume~80 of {\em EPiC Series in Computing}, pages 133--152. EasyChair, 2021.
\newblock \href {https://doi.org/10.29007/xwl1} {\path{doi:10.29007/xwl1}}.

\bibitem{ARCH22}
G.~Ernst et~al.
\newblock {ARCH-COMP} 2022 category report: {F}alsification with unbounded
  resources.
\newblock In G.~Frehse, M.~Althoff, E.~Schoitsch, and J.~Guiochet, editors,
  {\em 9th International Workshop on Applied Verification of Continuous and
  Hybrid Systems. {ARCH22}}, volume~90 of {\em EPiC Series in Computing}, pages
  203--221. EasyChair, 2022.
\newblock \href {https://doi.org/10.29007/fhnk} {\path{doi:10.29007/fhnk}}.

\bibitem{2021:falsification_of_hybrid_systems_using_adaptive}
G.~Ernst, S.~Sedwards, Z.~Zhang, and I.~Hasuo.
\newblock Falsification of hybrid systems using adaptive probabilistic search.
\newblock {\em ACM Trans. Model. Comput. Simul.}, 31(3), 2021.
\newblock \href {https://doi.org/10.1145/3459605} {\path{doi:10.1145/3459605}}.

\bibitem{2009:robustness_of_temporal_logic_specifications_for_continuous}
G.~Fainekos and G.~J. Pappas.
\newblock Robustness of temporal logic specifications for continuous-time
  signals.
\newblock {\em Theor. Comput. Sci.}, 410:4262--4291, 2009.
\newblock \href {https://doi.org/10.1016/j.tcs.2009.06.021}
  {\path{doi:10.1016/j.tcs.2009.06.021}}.

\bibitem{2023:search_based_software_testing_driven}
F.~Formica, T.~Fan, and C.~Menghi.
\newblock Search-based software testing driven by automatically generated and
  manually defined fitness functions.
\newblock {\em ACM Trans. Softw. Eng. Methodol.}, 33(2), 2023.
\newblock \href {https://doi.org/10.1145/3624745} {\path{doi:10.1145/3624745}}.

\bibitem{SBST-toolcomp22}
A.~Gambi, G.~Jahangirova, V.~Riccio, and F.~Zampetti.
\newblock {SBST} tool competition 2022.
\newblock In {\em 15th {IEEE/ACM} International Workshop on Search-Based
  Software Testing. {SBST} 2022}, pages 25--32. {IEEE}, 2022.
\newblock \href {https://doi.org/10.1145/3526072.3527538}
  {\path{doi:10.1145/3526072.3527538}}.

\bibitem{wgan_gp}
I.~Gulrajani, F.~Ahmed, M.~Arjovsky, V.~Dumoulin, and A.~Courville.
\newblock Improved training of {W}asserstein {GAN}s.
\newblock In {\em Proceedings of the 31st International Conference on Neural
  Information Processing Systems}, NIPS'17, pages 5769--5779. Curran Associates
  Inc., 2017.

\bibitem{eosl}
T.~Hastie, R.~Tibshirani, and J.~Friedman.
\newblock {\em Elements of Statistical Learning}.
\newblock Springer Series in Statistics. Springer, 2009.
\newblock 2nd edition.

\bibitem{2018:verification_challenges_in_f16_ground_collision_avoidance}
P.~Heidlauf, A.~Collins, M.~Bolender, and S.~Bak.
\newblock Verification challenges in {F}-16 ground collision avoidance and
  other automated maneuvers.
\newblock In G.~Frehse, editor, {\em 5th International Workshop on Applied
  Verification of Continuous and Hybrid Systems. {ARCH18}}, volume~54 of {\em
  EPiC Series in Computing}, pages 208--217. EasyChair, 2018.
\newblock \href {https://doi.org/10.29007/91x9} {\path{doi:10.29007/91x9}}.

\bibitem{2015:benchmarks_for_temporal_logic_requirements_for_automotive}
B.~Hoxha, H.~Abbas, and G.~Fainekos.
\newblock Benchmarks for temporal logic requirements for automotive systems.
\newblock In G.~Frehse and M.~Althoff, editors, {\em ARCH14--15. 1st and 2nd
  International Workshop on Applied veRification for Continuous and Hybrid
  Systems}, pages 25--30. EasyChair, 2015.
\newblock \href {https://doi.org/10.29007/xwrs} {\path{doi:10.29007/xwrs}}.

\bibitem{2022:ambiegen_tool_at_the_sbst_2022_tool_competition}
D.~Humeniuk, G.~Antoniol, and F.~Khomh.
\newblock Ambie{G}en tool at the {SBST} 2022 tool competition.
\newblock In {\em 15th {IEEE/ACM} International Workshop on Search-Based
  Software Testing. {SBST} 2022}, pages 43--46, 2022.
\newblock \href {https://doi.org/10.1145/3526072.3527531}
  {\path{doi:10.1145/3526072.3527531}}.

\bibitem{2022:a_search_based_framework_for_automatic_generation}
D.~Humeniuk, F.~Khomh, and G.~Antoniol.
\newblock A search-based framework for automatic generation of testing
  environments for cyber–physical systems.
\newblock {\em Inf. Softw. Technol.}, 149, 2022.
\newblock \href {https://doi.org/10.1016/j.infsof.2022.106936}
  {\path{doi:10.1016/j.infsof.2022.106936}}.

\bibitem{2023:ambiegen_a_search_based_framework_for_autonomous}
D.~Humeniuk, F.~Khomh, and G.~Antoniol.
\newblock Ambie{G}en: {A} search-based framework for autonomous systems
  testing.
\newblock {\em Sci. Comput. Program.}, 230, 2023.
\newblock \href {https://doi.org/10.1016/j.scico.2023.102990}
  {\path{doi:10.1016/j.scico.2023.102990}}.

\bibitem{2024:reinforcement_learning_informed_evolutionary_search}
D.~Humeniuk, F.~Khomh, and G.~Antoniol.
\newblock Reinforcement learning informed evolutionary search for autonomous
  systems testing.
\newblock {\em ACM Trans. Softw. Eng. Methodol.}, 2024.
\newblock \href {https://doi.org/10.1145/3680468} {\path{doi:10.1145/3680468}}.

\bibitem{SBFT-toolcomp24-uav}
S.~Khatiri, P.~Saurabh, T.~Zimmermann, C.~Munasinghe, C.~Birchler, and
  S.~Panichella.
\newblock {SBFT} tool competition 2024 - {CPS-UAV} test case generation track.
\newblock In {\em 17th {ACM/IEEE} International Workshop on Search-Based and
  Fuzz Testing. {SBFT} 2024}, pages 29–--32. ACM, 2024.
\newblock \href {https://doi.org/10.1145/3643659.3643931}
  {\path{doi:10.1145/3643659.3643931}}.

\bibitem{2002:safety_critical_sytems_challenges_and_directions}
J.~C. Knight.
\newblock Safety critical systems: {C}hallenges and directions.
\newblock In W.~Tracz, M.~Young, and Jeff Magee, editors, {\em Proceedings of
  the 24th International Conference on Software Engineering. {ICSE} 2002},
  pages 547--550. {ACM}, 2002.
\newblock \href {https://doi.org/10.1145/581339.581406}
  {\path{doi:10.1145/581339.581406}}.

\bibitem{2012:efficient_backprop}
Y.~A. LeCun, L.~Bottou, G.~B. Orr, and KR. M\"uller.
\newblock Efficient backprop.
\newblock In {\em Neural Networks: Tricks of the Trade}, volume 7700 of {\em
  Lecture Notes in Computer Science}, pages 9--48. Springer, 2012.
\newblock \href {https://doi.org/10.1007/978-3-642-35289-8_3}
  {\path{doi:10.1007/978-3-642-35289-8_3}}.

\bibitem{maler2004monitoring}
O.~Maler and D.~Nickovic.
\newblock Monitoring temporal properties of continuous signals.
\newblock In Y.~Lakhneck and S.~Yovine, editors, {\em Formal Techniques,
  Modelling and Analysis of Timed and Fault-Tolerant Systems. FORMATS 2004},
  volume 3253 of {\em Lecture Notes in Computer Science}, pages 152--166.
  Springer, 2004.
\newblock \href {https://doi.org/10.1007/978-3-540-30206-3_12}
  {\path{doi:10.1007/978-3-540-30206-3_12}}.

\bibitem{2021:efficient_optimization_based_falsification_of_cyber}
L.~Mathesen, G.~Pedrielli, and G.~Fainekos.
\newblock Efficient optimization-based falsification of cyber-physical systems
  with multiple conjunctive requirements.
\newblock In {\em IEEE 17th International Conference on Automation Science and
  Engineering. {CASE} 2021}, pages 732--737, 2021.
\newblock \href {https://doi.org/10.1109/CASE49439.2021.9551474}
  {\path{doi:10.1109/CASE49439.2021.9551474}}.

\bibitem{2021:stochastic_optimization_with_adaptive_restart_a_framework}
L.~Mathesen, G.~Pedrielli, S.~H. Ng, and Z.~B. Zabinsky.
\newblock Stochastic optimization with adaptive restart: A framework for
  integrated local and global learning.
\newblock {\em J. Global Optim.}, 79:87--110, 2021.
\newblock \href {https://doi.org/10.1007/s10898-020-00937-5}
  {\path{doi:10.1007/s10898-020-00937-5}}.

\bibitem{ARCH23}
C.~Menghi et~al.
\newblock {ARCH-COMP} 2023 category report: {F}alsification.
\newblock In G.~Frehse and M.~Althoff, editors, {\em 10th International
  Workshop on Applied Verification of Continuous and Hybrid Systems. {ARCH23}},
  volume~96 of {\em EPiC Series in Computing}, pages 151--169, 2023.
\newblock \href {https://doi.org/10.29007/6nqs} {\path{doi:10.29007/6nqs}}.

\bibitem{2020:approximation_refinement_testing_of_compute_intensive}
C.~Menghi, S.~Nejati, L.~C. Briand, and Y.~Isasi Parache.
\newblock Approximation-refinement testing of compute-intensive cyber-physical
  models: {A}n approach based on system identification.
\newblock In G.~Rothermel and D.{-}H. Bae, editors, {\em 42nd International
  Conference on Software Engineering. {ICSE2020}}, pages 372--384. {ACM}, 2020.
\newblock \href {https://doi.org/10.1145/3377811.3380370}
  {\path{doi:10.1145/3377811.3380370}}.

\bibitem{2017:unrolled_generative_adversarial_networks}
L.~Metz, B.~Poole, D.~Pfau, and J.~Sohl-Dickstein.
\newblock Unrolled generative adversarial networks.
\newblock 2017.
\newblock Preprint.
\newblock \href {https://arxiv.org/abs/1611.02163} {\path{arXiv:1611.02163}}.

\bibitem{2021:salvo_automated_generation_of_diversified_tests}
V.~Nguyen, S.~Huber, and A.~Gambi.
\newblock {SALVO}: {A}utomated generation of diversified tests for self-driving
  cars from existing maps.
\newblock In {\em 2021 IEEE International Conference on Artificial Intelligence
  Testing (AITest)}, pages 128--135, 2021.
\newblock \href {https://doi.org/10.1109/AITEST52744.2021.00033}
  {\path{doi:10.1109/AITEST52744.2021.00033}}.

\bibitem{SBST-toolcomp21}
S.~Panichella, A.~Gambi, F.~Zampetti, and V.~Riccio.
\newblock {SBST} tool competition 2021.
\newblock In {\em IEEE/ACM 14th International Workshop on Search-Based Software
  Testing. {SBST} 2021}, pages 20--27. {IEEE}, 2021.
\newblock \href {https://doi.org/10.1109/SBST52555.2021.00011}
  {\path{doi:10.1109/SBST52555.2021.00011}}.

\bibitem{2021:part_x_a_family_of_stochastic_algorithms}
G.~Pedrielli et~al.
\newblock Part-{X}: {A} family of stochastic algorithms for search-based test
  generation with probabilistic guarantees.
\newblock {\em IEEE Transactions on Automation Science and Engineering}, pages
  1--22.
\newblock \href {https://doi.org/10.1109/TASE.2023.3297984}
  {\path{doi:10.1109/TASE.2023.3297984}}.

\bibitem{ogan_full}
J.~{Peltom\"{a}ki} and I.~{Porres}.
\newblock Requirement falsification for cyber-physical systems using generative
  models.
\newblock 2023.
\newblock Preprint.
\newblock \href {https://arxiv.org/abs/2310.20493} {\path{arXiv:2310.20493}}.

\bibitem{sbst_long}
J.~Peltom\"aki, F.~Spencer, and I.~Porres.
\newblock Wasserstein generative adversarial networks for online test
  generation for cyber physical systems.
\newblock In {\em 15th {IEEE/ACM} International Workshop on Search-Based
  Software Testing. {SBST} 2022}, pages 1--5. {IEEE}, 2022.
\newblock \href {https://doi.org/10.1145/3526072.3527522}
  {\path{doi:10.1145/3526072.3527522}}.

\bibitem{sbst_short}
J.~Peltom\"aki, F.~Spencer, and I.~Porres.
\newblock {WOGAN} at the {SBST} 2022 {CPS} tool competition.
\newblock In {\em 15th {IEEE/ACM} International Workshop on Search-Based
  Software Testing. {SBST} 2022}, pages 53--54. {IEEE}, 2022.
\newblock \href {https://doi.org/10.1145/3526072.3527535}
  {\path{doi:10.1145/3526072.3527535}}.

\bibitem{2022:falsification_of_multiple_requirements_for_cyber_physical}
J.~Peltomäki and I.~Porres.
\newblock Falsification of multiple requirements for cyber-physical systems
  using online generative adversarial networks and multi-armed bandits, 2022.
\newblock \href {https://doi.org/10.1109/ICSTW55395.2022.00018}
  {\path{doi:10.1109/ICSTW55395.2022.00018}}.

\bibitem{2024:testing_cyber_physical_systems_with_explicit}
J.~Peltomäki, J.~Winsten, M.~Methais, and I.~Porres.
\newblock Testing cyber-physical systems with explicit output coverage.
\newblock In {\em 8th International Workshop on Testing Extra-Functional
  Properties and Quality Characteristics of Software Systems (ITEQS)}, 2024.
\newblock To appear.

\bibitem{ogan}
I.~Porres, H.~Rexha, and S.~Lafond.
\newblock Online {GAN}s for automatic performance testing.
\newblock In {\em IEEE International Conference on Software Testing,
  Verification and Validation Workshops. {ICSTW} 2021}, pages 95--100, 2021.
\newblock \href {https://doi.org/10.1109/ICSTW52544.2021.00027}
  {\path{doi:10.1109/ICSTW52544.2021.00027}}.

\bibitem{2020:model_based_exploration_of_the_frontier_of_behaviours}
V.~Riccio and P.~Tonella.
\newblock Model-based exploration of the frontier of behaviours for deep
  learning system testing.
\newblock In {\em Proceedings of the 28th ACM Joint Meeting on European
  Software Engineering Conference and Symposium on the Foundations of Software
  Engineering}, pages 876--888. ACM, 2020.
\newblock \href {https://doi.org/10.1145/3368089.3409730}
  {\path{doi:10.1145/3368089.3409730}}.

\bibitem{2011:a_survey_of_cyber_physical_systems}
J.~Shi, J.~Wan, H.~Yan, and H.~Suo.
\newblock A survey of cyber-physical systems.
\newblock In {\em 2011 International Conference on Wireless Communications {\&}
  Signal Processing. {WCSP} 2011}, pages 1--6. {IEEE}, 2011.
\newblock \href {https://doi.org/10.1109/WCSP.2011.6096958}
  {\path{doi:10.1109/WCSP.2011.6096958}}.

\bibitem{2021:psy_taliro_a_python_toolbox_for_search}
Q.~Thibeault, J.~Anderson, A.~Chandratre, G.~Pedrielli, and G.~Fainekos.
\newblock {PSY}-{T}a{L}i{R}o: A {P}ython toolbox for search-based test
  generation for cyber-physical systems.
\newblock In A.~L. Lafuente and A.~Mavridou, editors, {\em Formal Methods for
  Industrial Critical Systems. {FMICS} 2021}, volume 12863 of {\em Lecture
  Notes in Computer Science}. Springer, 2021.
\newblock \href {https://doi.org/10.1007/978-3-030-85248-1_15}
  {\path{doi:10.1007/978-3-030-85248-1_15}}.

\bibitem{2020:falsification_of_cyber_physical_systems_with_robustness}
M.~Waga.
\newblock Falsification of cyber-physical systems with robustness-guided
  black-box checking.
\newblock In {\em Proceedings of the 23rd International Conference on Hybrid
  Systems: Computation and Control}. ACM, 2020.
\newblock \href {https://doi.org/10.1145/3365365.3382193}
  {\path{doi:10.1145/3365365.3382193}}.

\bibitem{wogan_sbft23}
J.~Winsten and I.~Porres.
\newblock {WOGAN} at the {SBFT} 2023 tool competition - cyber-physical systems
  track.
\newblock In {\em 2023 {IEEE/ACM} International Workshop on Search-Based and
  Fuzz Testing. {SBFT} 2023}, pages 43--44. {IEEE}, 2023.
\newblock \href {https://doi.org/10.1109/SBFT59156.2023.00009}
  {\path{doi:10.1109/SBFT59156.2023.00009}}.

\bibitem{wogan_uav}
J.~Winsten, V.~Soloviev, J.~Peltom\"aki, and I.~Porres.
\newblock Adaptive test generation for unmanned aerial vehicles using
  {WOGAN-UAV}.
\newblock In {\em Proceedings of the 17th ACM/IEEE International Workshop on
  Search-Based and Fuzz Testing}, page 43–44. ACM, 2024.
\newblock \href {https://doi.org/10.1145/3643659.3648603}
  {\path{doi:10.1145/3643659.3648603}}.

\bibitem{2021:deephyperion_exploring_the_feature_space_of_deep}
T.~Zohdinasab, V.~Riccio, A.~Gambi, and P.~Tonella.
\newblock Deep{H}yperion: exploring the feature space of deep learning-based
  systems through illumination search.
\newblock In {\em Proceedings of the 30th ACM SIGSOFT International Symposium
  on Software Testing and Analysis}, pages 79--90. ACM, 2021.
\newblock \href {https://doi.org/10.1145/3460319.3464811}
  {\path{doi:10.1145/3460319.3464811}}.

\end{thebibliography}
\bibliographystyle{plainurl}

\end{document}